\documentclass{article}
\usepackage{algorithm}
\usepackage[noend]{algpseudocode}
\usepackage{mathtools}
\usepackage{amsthm}
\usepackage{array}
\usepackage{rotating,tabularx}
\usepackage{booktabs,caption}
\usepackage{multirow}
\usepackage[flushleft]{threeparttable}
\usepackage[utf8]{inputenc}
\usepackage{amssymb,amsmath}
\usepackage[numbers]{natbib}
\usepackage{graphicx}
\graphicspath{images/}
\usepackage{comment}
\usepackage{hyperref}
\usepackage{xcolor}
\usepackage{authblk}
\newtheorem{definition}{Definition}
\newtheorem{theorem}{Theorem}
\newtheorem{lemma}{Lemma}

\title{Enforcing the Principle of Locality for Physical Simulations with Neural Operators}

\author[1]{Jiangce Chen}
\author[1]{Wenzhuo Xu}
\author[1]{Zeda Xu}
\author[1]{Noelia Grande Guti\'{e}rrez}
\author[1]{Sneha Prabha Narra}
\author[1]{Christopher McComb\thanks{ccm@cmu.edu \vspace{-1em}Address all correspondence to this author.}}
\affil[1]{Carnegie Mellon University\\
	Pittsburgh, PA, USA}

\begin{document}
\maketitle

\begin{abstract}
Time-dependent partial differential equations (PDEs) for classic physical systems are established based on the conservation of mass, momentum, and energy, which are ubiquitous in scientific and engineering applications. These PDEs are strictly local-dependent according to the principle of locality in physics, which means that the evolution at a point is only influenced by the neighborhood around it whose size is determined by the length of timestep multiplied with the speed of characteristic information traveling in the system. However, deep learning architecture cannot strictly enforce the local-dependency as it inevitably increases the scope of information to make local predictions as the number of layers increases. Under limited training data, the extra irrelevant information results in sluggish convergence and compromised generalizability. This paper aims to solve this problem by proposing a data decomposition method to strictly limit the scope of information for neural operators making local predictions, which is called data decomposition enforcing local-dependency (DDELD).
The numerical experiments over multiple physical phenomena show that DDELD significantly accelerates training convergence and reduces test errors of benchmark models on large-scale engineering simulations. 

\end{abstract}

\section{Introduction}
\begin{figure}[h]
    \centering
    \includegraphics[width=0.6\linewidth]{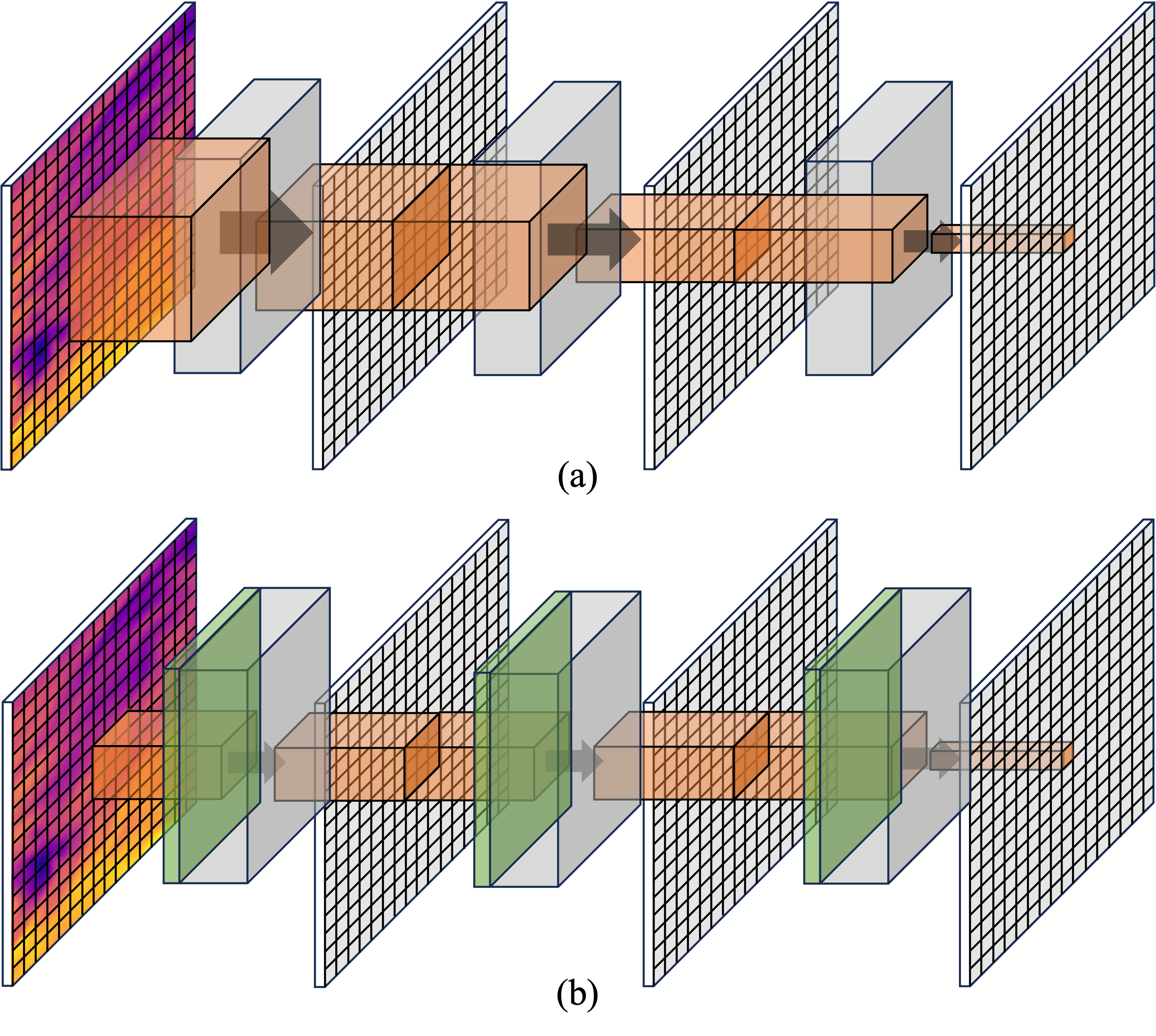}
    \caption{The target problem. (a) The deep learning architecture inevitably expands the scope of input data used for the prediction at one position as the number of layers increases, which is not compatible with the local-dependency assumption for a classical physics system. (b) DDELD method proposed in this paper can ensure that the scope of the input data stays constant regardless of the number of layers, which decouples the expressiveness and local dependency of neural networks.}
    \label{fig:problem}
\end{figure}
A broad variety of classical, time-dependent physical systems involve the evolution of mass, momentum, and energy. These relationships can be described by partial differential equations (PDEs) with local dependency under the assumption that the physical information travels at a limited speed through the physical medium. Solving these PDEs are fundamental to overcoming engineering challenges like airplane design \cite{martins2022aerodynamic,sugaya2022turbulent}, additive manufacturing control \cite{chen2024accelerating},  weather forecasting \cite{pathak2022fourcastnet,lam2022graphcast}, drug delivery \cite{boso2020drug}, and pandemic outbreak modeling \cite{raheja2023machine}. Traditional methods for solving these PDEs in discretized form include finite difference methods, finite element methods, and finite volume methods. However, all of these incurs a cubic relationship between domain resolution and computation cost \cite{kochkov2021machine}, which means that a 10-fold increase in resolution leads to a 1000-fold increase for 3D problem in the computational cost. 
Advancements in computation infrastructure and parallel computing have paved the way for the success of machine learning (ML). 
This, in turn, signals a paradigm shift in scientific computation, with ML techniques emerging as valuable tools for addressing the computational limitations with sub-cubic costs \cite{vinuesa2022enhancing,chen2023capturing}.

\paragraph{State-of-the-art methods} Physics-informed neural networks (PINNs) have demonstrated the ability to learn the smooth solutions of known nonlinear PDEs with little or no data by incorporating PDE residuals into the training loss \cite{raissi2019physics}. This approach has proven particularly valuable in solving inverse problems, enabling the identification of unknown coefficients in governing equations \cite{jagtap2020conservative}. However, it is important to note that a single PINN model is typically trained to learn one specific instance of a PDE with specific coefficients, initial conditions (IC) and boundary conditions (BC) \cite{kovachki2021neural}. Consequently, it is necessary to retrain the PINN model for every new PDE instance, and the associated large training cost stands out as a major limitation of the PINN framework  \cite{jagtap2021extended}. This limitation poses a challenge to the generalizability and efficiency of PINNs, hindering their ability to overcome computational limitations posed by traditional numerical methods. 

In addition, neural operators have emerged as an effective way to overcome computation bottlenecks in approximating the solutions for a family of PDEs by learning the mapping between function spaces from data, such as DeepONet \cite{lu2019deeponet} and Fourier Neural Operators (FNOs) \cite{li2020fourier}.
Besides, conservative laws incorporated into neural operators have been found can improve learning process with limited data \cite{liu2023harnessing, hansen2023learning,mouli2024using,lorin2024non,yang2024pde}.
The expressiveness and nonlinearity of these models are realized through deep learning architecture, iterative forward computations across multiple linear operators. 
However, the multiple-layer architecture of deep learning applied in these models cannot focus on the local evolution patterns for time-dependent PDEs as elaborated below.

\paragraph{Unsolved problem}{
In the context of classical systems, information propagates at a limited speed. This implies that the physics properties at a position in the next time step depend on the current status of its neighbors, which is called local-dependency. An ML model for time-dependent PDEs can be formulated as a neural operator, mapping the current status of the system into the status at the next time step. So a neural operator approximating a family of time-dependent PDEs in classical systems should also have local-dependent property, which only utilizes a specified window of information around a position to make the local prediction.
Even though some neural networks, such as CNN based methods \cite{he2016deep,wang2020towards}, Graph-CNN based methods \cite{pfaff2020learning,li2020neural}, and neural operator with localized kernels \cite{liu2024neural}, have the property of local-dependency in one layer, the multiple-layer architecture will expand the size of information window unintentionally as the layer number increases as illustrated in Figure \ref{fig:problem} (a) and will be further discussed in Section \ref{sec:imcompatibility}. Given a specific time step, the local information near the boundary of the enlarged window would be beyond its reach to have causal effects on the center of the window. Therefore, it only adds noise, which distracts the neural operator from capturing the true local physical patterns over the center of the window.

\paragraph{Contributions} { 
This paper introduces a data decomposition method for neural operators, called data decomposition enforcing local-dependency (DDELD), to ensure strict local-dependency in making the predictions for time-dependent PDEs of classic physical systems, as illustrated in Figure \ref{fig:problem}. Our major contributions can be summarized as follows.
\begin{itemize}
    \item We demonstrate that the multiple-layer architecture is not suitable for local-dependency PDEs as demonstrated in Figure \ref{fig:problem} (a).
    \item We establish a method to solve the incompatibility. It decomposes the domain into small windows based on local-dependency, and integrates these windows with linear time complexity, which underscores its efficiency and scalability across different problem sizes.
    \item We apply DDELD to the benchmark neural operators over the data of complex fluid mechanics. Our results demonstrate a significant enhancement in the convergence rate and generalization capabilities of benchmark neural operators. 
\end{itemize}}

\paragraph{Remarks}{The method proposed in this paper shares similarities with domain decomposition methods commonly used in traditional numerical approaches. Domain decomposition methods aim to alleviate the hardware demands of solving a large domain by partitioning it into smaller regions that can be solved in parallel with defined interface conditions \cite{haddadi2017cost,tang2021review,calvo2024robust,taddei2024non}. A series of ML models have incorporated the concept of domain decomposition \cite{li2019d3m,li2020deep,jagtap2021extended,pan2024domain}. These models decompose the domain into subdomains, each assigned a PINN model that is trained independently. Information exchange between subdomains is facilitated by adjusting the boundary term or incorporating interface conditions in the training loss. Although these methods enhance computation speed by solving subdomains in parallel, they share limitations inherent in PINNs—specifically, they are tailored to a particular instance of a PDE and involve a time-consuming training process. In contrast, DDELD proposed in this paper is designed for neural operators that approximate the solutions of a family of PDEs while also supports parallel computation.}

\section{Background of Neural Operators}
PDEs can be viewed as nonlinear operators that map between Banach function spaces.
We formulate the ML models approximating a family of time-dependent PDEs as the nonlinear operators following the work of Li \textit{et. al.}\cite{li2020fourier}. 

\paragraph{Neural operator formulation}
The domain of a classical physical system in $d$-dimensional space is denoted as $D \subset \mathbb{R}^d$ which is a bounded open set. Let $d_u$ be the dimension of the physical properties evolved in the system. Let $d_a$ be the dimension of the constant properties of a specific instance of PDEs, such as coefficients. Let $\mathcal{A}=\mathcal{A}(D; \mathbb{R}^{d_a})$ and $\mathcal{U}=\mathcal{U}(D; \mathbb{R}^{d_u})$ be Banach spaces of functions that take values in $\mathbb{R}^{d_a}$ and $\mathbb{R}^{d_u}$, respectively. The constant properties of the system are denoted as $a \in \mathcal{A}$. The status of the system at time $t$ is denoted as $u^t \in \mathcal{U}$. For the convenience of formulation, the time dimension is discretized uniformly. We have $t=0,1,2,..., T$ with fixed timestep $\Delta t$ and maximum $T$. The evolution of the system from $t$ to $t+1$ can then be represented by a nonlinear operator $G^{\dag}: \mathcal{A} \times \mathcal{U} \rightarrow \mathcal{U}$ in the way that
\begin{equation}
    u^{t+1} = G^{\dag}(a,u^t).
\end{equation}
Given $a$ and the initial status of the system $u^0$, $u^t$ can be calculated by $G^{\dag}$ in an iterative way for all $t=0,1,2,...$. So, $G^{\dag}$ can be viewed as the solution operator of a family of time-dependent PDEs characterized by $\mathcal{A}$. 

\paragraph{Learning framework}{Given $a_j$, the solution of the instance of PDEs specified by $a_j$ is the list $[u^0_j, u^1_j,...,u^T_j]$ which is denoted as $U_j$. Suppose we have observations $\{a_j,U_j\}_{j=1}^N$ where $a_j \thicksim \mu$ is an i.i.d. sequence from the probability measure $\mu$ supported on $\mathcal{A}$, and $U_j$ is the corresponding solution of the PDEs specified by $a_j$. Our goal is to approximate $G^{\dag}$ by constructing a parametric non-linear operator, named neural operator, $G_{\theta}: \mathcal{A}\times \mathcal{U} \rightarrow \mathcal{U}$, where $\theta \in \Theta$ denotes the set of neural network parameters. With a cost function $C: \mathcal{U}\times \mathcal{U} \rightarrow \mathbb{R}$, the neural operator is trained by the following learning framework
\begin{equation}
\min_{\theta \in \Theta}\mathbb{E}_{a_j \sim \mu}[ \mathbb{E}_{u^t_j \sim U_j}[ C(G_{\theta}(a_j,u^t_j),u^{t+1}_j) ]].
\end{equation}}

\paragraph{Discretization}{
The functions $a_j$ and $u_j$ are discretized for the ease of computation in practice. According to the definition in \cite{kovachki2021neural}, the neural operator is mesh-independent, which means that the same $G_{\theta}$ can be used for different discretizations and $C(G_{\theta}(a_j,u^t_j),u^{t+1}_j)$ should not have significant changes over different discretizations. In this paper, we limit the discussion to the same discretization, so we relax the mesh-independent requirements. We generalized the definition of the neural operator to any data-driven models that can take $a_j$ and $u^t_j$, output $u^{t+1}_j$ over the fixed discretization.}

\section{Incompatibility between Deep Learning and Local-dependency}
\subsection{Local-dependency}
The physical information in a classical physical (non-quantum mechanics) system travels at a limited speed. In classical physics, the principle of locality states \cite{bell1964einstein} that for a cause at one point to have an effect at another point, something in the space between those points must mediate the action. To exert an influence, something, such as a wave or particle, must travel through the space between the two points, carrying the influence.  The special theory of relativity limits the maximum speed at which causal influence can travel to the speed of light. In a regular engineering system that involves the evolution of the distribution of mass, momentum, and energy, the physical information is usually driven by the effects of fluctuation, diffusion, and convection, whose speed is much slower than light. For example, in linear advection system, the information traveling speed can be characterized by the velocity field, and the information near the local up-winding region is preferred to take into account for evolution simulation. In this paper, we aims to develop a general method to all the time-dependent classical physical systems that the local-dependency can be strictly applied for ML models. 
\begin{definition}
Let $\delta$ be the maximal length that the physical information can travel in $\Delta t$. So, the physical properties at a point $x\in D$ can only be influenced by its neighborhood $U(x,\delta)=\{y|  y\in D, d(x,y) < \delta\}$ within $\Delta t$ timestep. Let $d(\cdot,\cdot)$ be the distance metric defined in $\mathbb{R}^d$. To predict $u^{t+1}(x)$, we do not need the whole system status $u^t$, but only the system status in $U(x,\delta)$ is sufficient. $U(x,\delta)$ is defined as the local-dependent region of the system at $x$.
\end{definition}

We define the segment of $u^t$ over $U(x,\delta)$ as
\begin{equation}\label{equ:segmentation}
    u^t|_{U(x,\delta)}:= \left \{ \begin{array}{rcl}
        u^t & \mbox{for} & x \in U(x,\delta) \\
        0 & \mbox{for} & x \notin  U(x,\delta) \\
    \end{array}\right \}
\end{equation}
So, we can define the local-dependent operator for the system in Definition \ref{def:local_operator}.
\begin{definition}\label{def:local_operator}
    A nonlinear operator $G^{\dag}: \mathcal{A} \times \mathcal{U} \rightarrow \mathcal{U}$  is said to have the local-dependency property and thus called a local-dependent operator if it updates the system as 
    $$
    u^{t+1}(x) = G^{\dag}(a|_{U(x,\delta)},u^t|_{U(x,\delta)}), \forall x \in D.
    $$
\end{definition}

\subsection{The more the layers, the weaker the local-dependency}\label{sec:imcompatibility}
Here we explain why the deep learning architecture of neural operators weakens local-dependency. 
A neural operator consists of multiple layers where each layer is a linear operator followed by a non-linear activation. The universal approximation theorem states that such architecture can accurately approximate any nonlinear operator \cite{chen1995universal}. 
The deep learning architecture of the neural operator for time-dependent PDEs can be formulated in an iterative way that
\begin{equation}\label{equ:updates}
\begin{aligned}
    & v_0 = P(a_j,u^t_j ) \\
    & v_{i+1} = \sigma( K_{\phi} (v_i)) \\
    & u^{t+1}_j = Q(v_m). \\
\end{aligned}
\end{equation}
At the beginning, the input $a_j$ and $u^t_j$ is concatenated and projected to a higher dimension space $\mathbb{R}^{d_v}$ using a local linear transformation $P: \mathbb{R}^{d_a+d_u} \rightarrow \mathbb{R}^{d_v}$. Next, a series of iterative updates are applied, generating $v_0 \mapsto v_1 ... \mapsto v_m$, where each vector takes value in $\mathbb{R}^{d_v}$. Finally, $v_m$ is projected back by a local linear transformation $Q: \mathbb{R}^{d_v} \rightarrow \mathbb{R}^{d_u}$. Let $\mathcal{V}=\mathcal{V}(D; \mathbb{R}^{d_v})$ be a Banach space of functions that take values in $\mathbb{R}^{d_v}$.
The iterative update consists of a parameterized linear operator $K_{\phi}: \mathcal{V} \rightarrow \mathcal{V}$ followed by a non-linear activation function $\sigma: \mathbb{R} \rightarrow \mathbb{R}$.

Common linear operators include graph-based operators \cite{li2020neural}, low-rank operators \cite{borm2004low}, multipole graph-based operators \cite{li2020multipole}, and Fourier operators \cite{kovachki2021neural}. 
It is possible to define a linear operator that only involves the local information around a point. For example, convolution is one of the common linear operators. We can define a local-dependent convolution over local-dependent region $U(x,\delta)$ as
\begin{equation}\label{equ:local_convolution}
    K_{\phi}(v_i)(x) = \int_{U(x,\delta)} k_{\phi}(x-y)v_i(y)dy, \forall x\in D,
\end{equation}
where $k_{\phi}$ is a family of parameterized periodic functions.
However, under a neural operator that consists of multiple layers of the local-dependent convolutions, the local-dependent region at $x$ is larger than $U(x,\delta)$. Specifically, the size of the expanded local-dependent region is positively proportional to the layer number, which is stated in Theorem \ref{theorem:local_dependent_region} and is proved in Appendix \ref{appendix:proof}.
\begin{theorem}\label{theorem:local_dependent_region}
Let $G_{\theta}: \mathcal{A}\times \mathcal{U} \rightarrow \mathcal{U}$ be a neural operator consisting of $k$ layers of local-dependent convolution defined in Equation \ref{equ:local_convolution} where the interval of the convolution is the $U(x, \delta)$. While the local-dependent region of each convolution layer is $U(x, \delta)$, the local-dependent region of the neural operator at $x$ is $U(x, k \delta)$.
\end{theorem}

This result presents the incompatibility. Under the deep learning architecture, to increase the expressiveness of neural network to account for the nonlinearity of the PDEs, we need to increase the layer number. However increasing the layer number results in expanding the scope of the information used to make the prediction, which might violate the local-dependency of time-dependent PDEs as defined in Definition \ref{def:local_operator}.

\section{Methodology}
\subsection{Method formulation}
Instead of limiting the scope of the linear operator to one layer, we propose limiting the scope of input data directly. 
DDELD decomposes the data so that each operator only works on the segmentation $u^t|_{U(x,\delta)}$ defined in Equation \ref{equ:segmentation}. So, now we have
\begin{equation}
\begin{aligned}
& v_0 = P(a_j,u^t_j|_{U(x,\delta)}) \\
& v_{i+1}|_{U(x,\delta)} = \sigma( K_{\phi} (v_i|_{U(x,\delta)})) \\
& u^{t+1}_j|_{U(x,\delta)} = Q(v_m|_{U(x,\delta)})
\end{aligned}
\end{equation}
Under this formulation, the calculation of $u^{t+1}_j(x)$ only involves the information in $U(x,\delta)$ no matter the scope of the linear operator $K_{\phi}$. Note that the same $K_{\phi}$ is used for all segmentations.
To realize the segmentation $u^t|_{U(x,\delta)}$ efficiently, we developed DDELD to partition the data into windows with prescribed sizes and integrate the predictions over the individual windows into the whole domain.

Given a domain discretized by a grid, as illustrated in Figure \ref{fig:concept} (a), to predict the physical properties in the next timestep at one position (colored in black), it is assumed that the local region colored in grey contains sufficient information. So, instead of inputting the whole domain into the ML model, we should only input the relevant local region (colored in grey) to make the prediction (colored in orange). Our domain decomposition algorithm partitions the domain evenly into smaller windows and has the ML model make the predictions at the centers of the windows. The details of the domain decomposition and its reverse, window patching, algorithms are illustrated in Figure \ref{fig:data_decomposition} and explained in Section \ref{sec:data_decomposition}.  While one decomposition of the domain only generates the prediction at the center of the windows, as shown in Figure \ref{fig:concept} (b), we need decompositions over an expanded domain as illustrated in Figure \ref{fig:domain_expansion} and detailed in Section \ref{sec:domain_expasion} to make the complete prediction as indicated in Figure \ref{fig:concept} (c). The prediction integration algorithm illustrated in Figure \ref{fig:data_integration} and detailed in Section \ref{sec:prediction_integration} explains how to obtain the prediction over the complete domain. 

The window size $\delta$ is selected as a hyper-parameter in the scale of a characteristic length $L_c$ for specific PDEs. It's difficult to apply a generalized term regarding the choice of window size for all the PDEs, but we can still give suggestions on the choice of window size based on whether the PDE is convection or diffusion dominant, or if the derived physical property has a clear frequency character (for example, a peak in the energy spectrum). The details of window size determination are discussed in Section \ref{sec:window_size}.

\begin{figure}[h]
    \centering
    \includegraphics[width=0.9\linewidth]{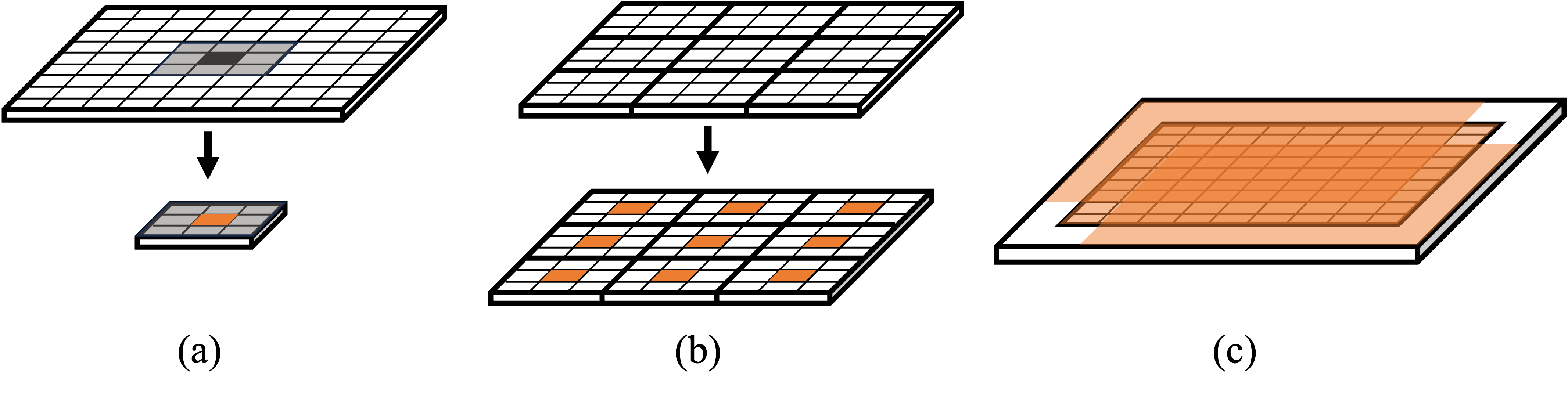}
    \caption{The overview of the method. (a) To predict the physical property at the black position, its neighbors (colored in grey) contain sufficient information. The prediction is colored in orange. (b) One decomposition of the domain can be used to make the predictions over a part of the domain. (c) Multiple decompositions and prediction integration algorithms are needed to make the prediction over the whole domain. }
    \label{fig:concept}
\end{figure}

\begin{figure}[h]
    \centering
    \includegraphics[width=0.6\linewidth]{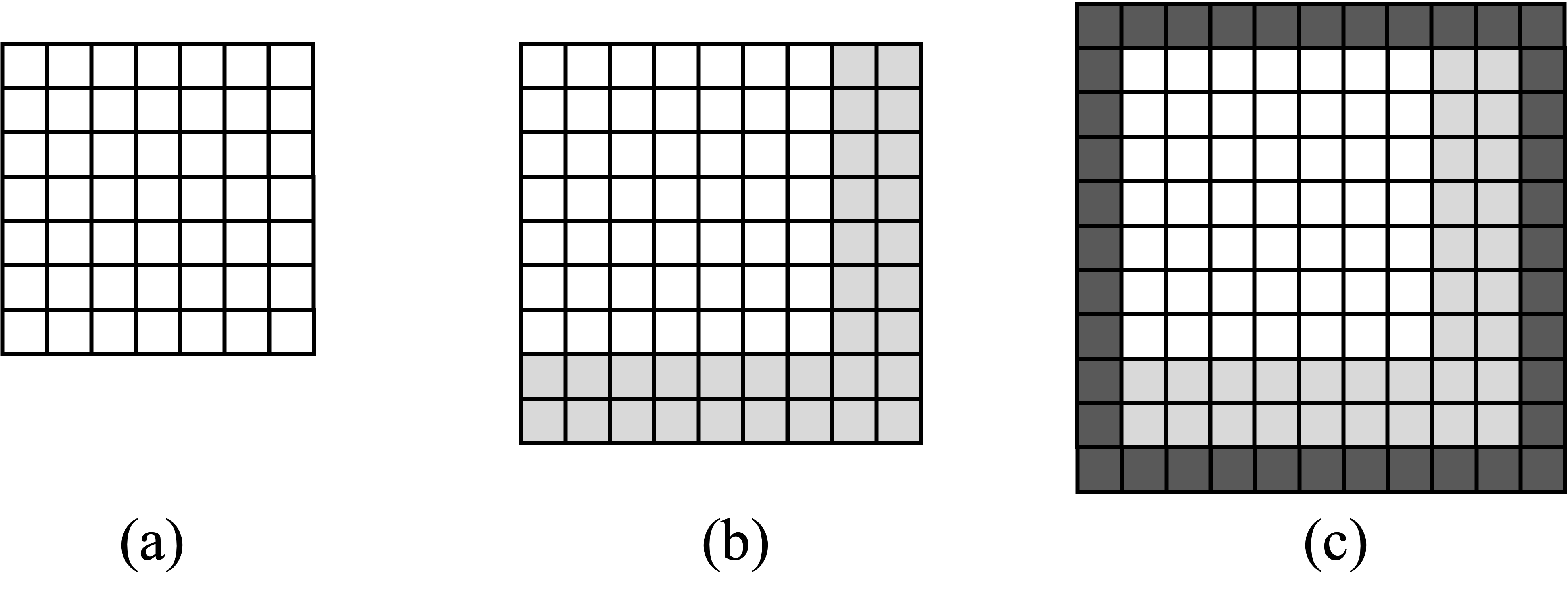}
    \caption{The example of expanding the domain in two steps. (a) Given window size $(3,3)$, a 2D domain with size $(7,7)$ needs to be expanded to be multiple of the window size. (b) In the first step, the domain is expanded to $(9,9)$ by padding the zeros at the end of each dimension. (c) In the second step, the domain is expanded to $(10,10)$ to be compatible with the prediction integration algorithm by padding zeros at the beginning and the end of each dimension. }
    \label{fig:domain_expansion}
\end{figure}

\begin{figure}[h]
    \centering
    \includegraphics[width=0.9\linewidth]{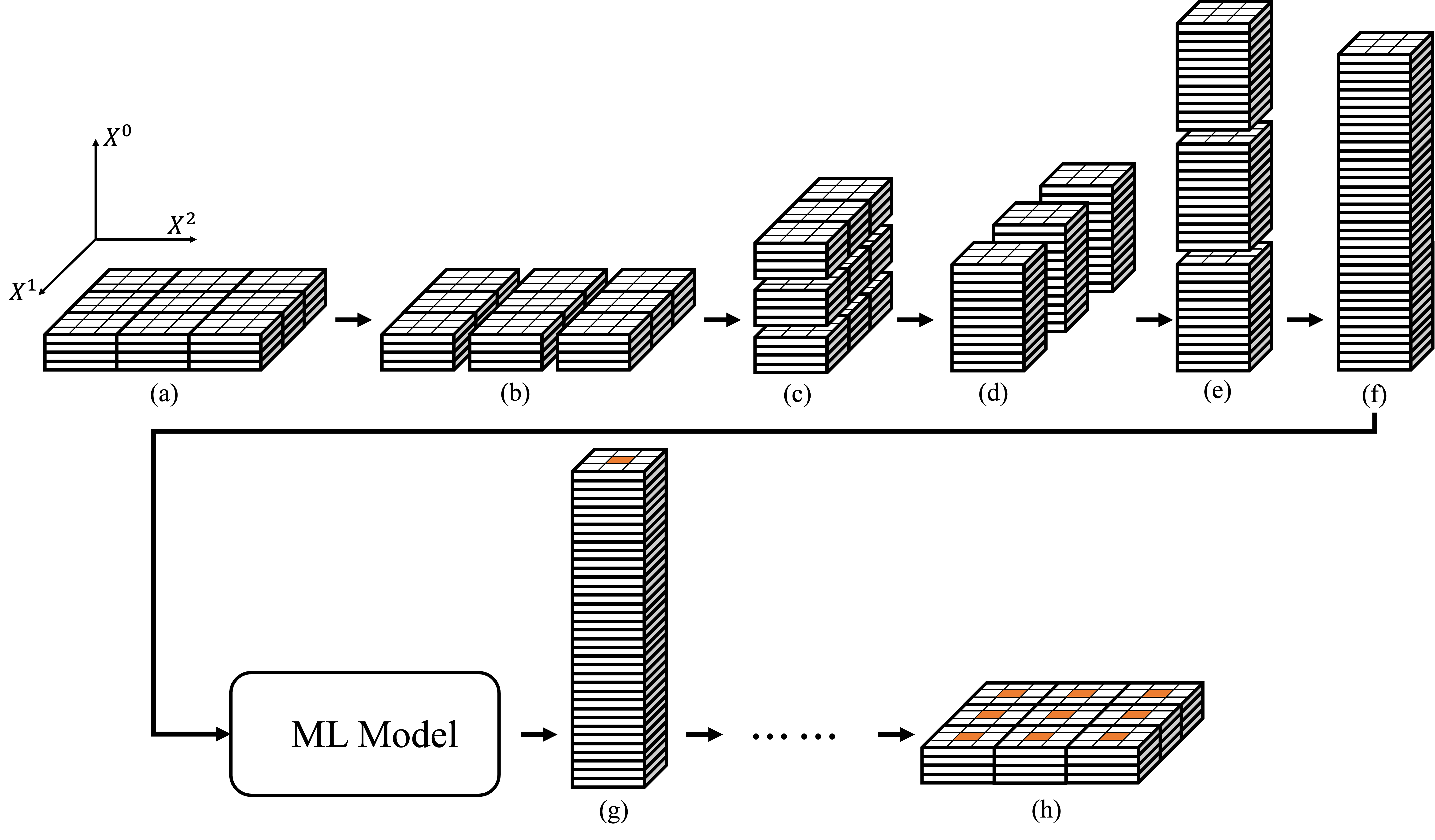}
    \caption{The illustration of the domain decomposition and window patching for one partition. (a). A data batch of global domain. In this example, the data in 2D space with $X^0$ denoting the batch dimension, $X^1$, and $X^2$ denoting the two domain dimensions.  The batch size is set as 4 and the domain is set to be decomposed into $3\times 3$ blocks in this example. (b) The batch is split into three parts in $x^2$ dimension. (c) The parts are stacked in $X^0$ dimensions to make a new batch with 12 batch sizes and $1\times 3$ blocks. (d) The batch is split into three parts in $x^1$ dimension. (e) The parts are stacked in $x^0$ dimension. (f) The original data is decomposed into $3\times 3$ blocks which are stacked to make a new batch with 36 batch size. (g) The ML model predicts the physical properties at the centers of the windows. (h) In a reverse of the decomposition process (b) to (e), the data shape is recovered to the original shape with the predictions made at the centers of each window.  }
    \label{fig:data_decomposition}
\end{figure}

\begin{figure}[h]
    \centering
    \includegraphics[width=0.9\linewidth]{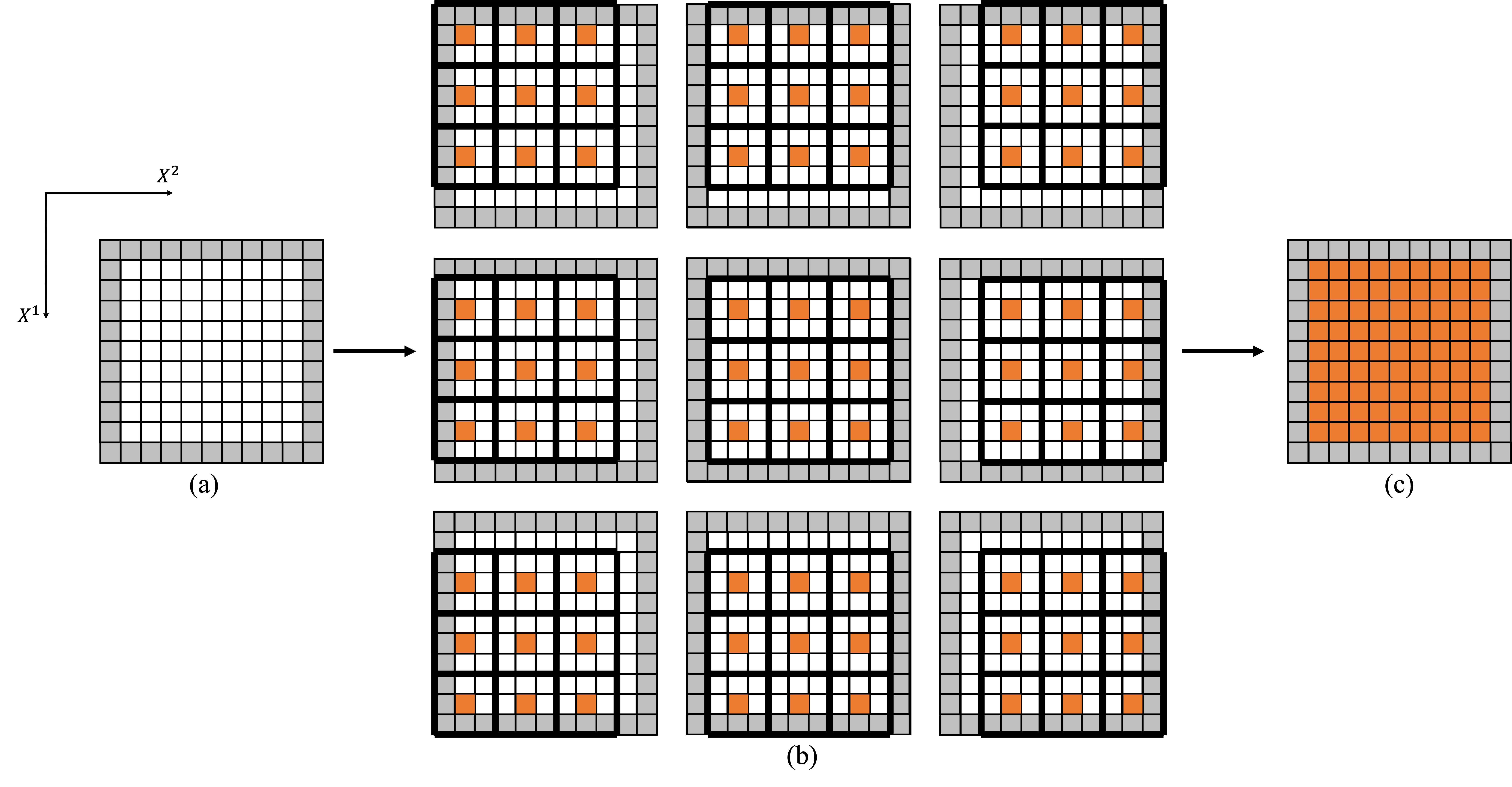}
    \caption{The illustration of the data integration algorithm. (a) A batch of 2D data is viewed from the top where $X^1$ and $X^2$ are the two domain dimensions and the batch size dimension $x^0$ is not shown in the top view. The domain represented by the grid is expanded by padding the zeros which is the blank space near its boundary. (b) Multiple partitions are made over the expanded domain. The predictions over the partitions are made independently and only the predictions in the centers of the blocks are preserved for the prediction integration, which is colored in orange. (c) The prediction over the whole domain is integrated.}
    \label{fig:data_integration}
\end{figure}

\subsection{Domain decomposition}\label{sec:data_decomposition}
Input data for the ML model is formed as batches. 
A batch of structured data can be represented as a tensor with size $(N_b, N_1,..,N_d,N_c)$ where $N_b$ is the batch number and $N_c$ is the channel dimension which is determined by the dimension of physical properties at a position. Given the window size $(W_1,...,W_d)$, our method aims to decompose the whole batch of data into a new tensor with size $(N_b*B_i*...*B_d,W_1,...,W_d,N_c)$. Algorithm \ref{alg:data_decomposition} shows the algorithm. Figure \ref{fig:data_decomposition} illustrates a 2D example. In this example, we are given a batch of 2D data with size $(4,9,9,1)$ and the window size is $(3,3)$. The block number is $(3,3)$. After a sequence of splitting and stacking operations, the batch of the whole domain data is converted to a batch of the windows in the shape of $(36,3,3,1)$. For 2D data, there is twice splitting and twice stacking, while for 3D data, there is thrice splitting and thrice stacking. The time complexity of the splitting and stacking of the array data is $O(B_{max})$ where $B_{max}$ is the maximal among $B_i,i=1,...,d$. The batch of the windows is then input into an ML model and the physical properties at the center of the windows are predicted as shown in Figure \ref{fig:data_decomposition} (g). Then the window patching algorithm detailed in Algorithm \ref{alg:window_patching}, a reverse of the decomposition operation, is followed to recover the batch of the whole domain from the batch of the windows. The window patching algorithm consists of the same number of splitting and stacking operations as the decomposition algorithm does, whose time complexity is also $O(B_{max})$. So the time complexity of the total algorithm is $O(B_{max})$.

\subsection{Domain expansion}\label{sec:domain_expasion}
A grid $d-$dimension domain with size $(N_1,...,N_d)$ where $N_i \in \mathbb{N}^{+}$ is denoted as $D(N_1,...,N_d)$ or $D$ if the dimensions are given in the context. 
To decompose the domain into the windows with size $(W_1,...,W_d)$ where $W_i \in \mathbb{N}^{+}$ and make it compatible with the data integration algorithm, the domain needs to be expanded by two steps. The first step is to pad the zeros after the end of each dimension to make the dimension number as the multiple of $W_i$. The second step is to pad $[w/2]$ zeros at the beginning and $[(w-1)/2]$ zeros after the end of each dimension. Figure \ref{fig:domain_expansion} illustrates the domain-expanding process in a 2D example. After the expansion, the new dimension number $N_i^{new}$ becomes
\begin{equation}
    N_i^{new} = ([(N_i-1)/W_i]+1)W_i + (W_i-1),\text{for }i=1,...,d,
\end{equation}
where $[\cdot]$ is the operation that only keeps the integer part of a real number.
In the following, the grid domain is always considered as the domain after expansion.
The number of the windows of the decomposition in each dimension is denoted as $B_i$, referring as the block number, can be calculated as
\begin{equation}
    B_i = [N_i/W_i].
\end{equation}

\subsection{Prediction integration}\label{sec:prediction_integration}
As shown in Figure \ref{fig:data_decomposition} (h) the prediction over one decomposition can only give us the physical properties at the centers of the windows, to get the complete prediction over the whole domain, we need multiple decompositions which ensures that all the positions are the centers of some windows. Algorithm \ref{alg:prediction_integration} details the prediction integration algorithm. Figure \ref{fig:data_integration} illustrates the prediction integration algorithm with a 2D example. Figure \ref{fig:data_integration} (a) shows the 2D grid domain. The blank zone near the boundary indicates the padding zeros. The original domain size is $(9,9)$ and the expanded domain size is $(11,11)$. With $(3,3)$ window size, the block number for one decomposition is $(3,3)$. As each decomposition can be used to predict the center of all its windows, we need $3\times 3$ different decompositions of the domain which can cover all the positions as shown in Figure \ref{fig:data_integration} (b) and (c). In general case, it is required to have $\prod_{i=1}^d W_i$ decompositions to cover the whole domain. The window size $W_i$ is a small number compared with the domain dimension. Since the predictions over the different decompositions are independent, they can be calculated in parallel. Therefore, the time complexity of the prediction integration algorithm is the constant multiple of the time complexity of the ML model inference. 

\begin{algorithm}[h]
\caption{Domain decomposition}
\begin{algorithmic}[1]
\Procedure{Chunk-domain}{$x, B, d$}       \Comment{domain tensor, block numbers, dimension number}
    \State $i \leftarrow 0$
    \While{$i \neq d$}  
        \State $x \leftarrow Split(x,B_i,i+1)$  \Comment{Split the x along $(i+1)$-th dimension into $B_i$ blocks.}
        \State $x \leftarrow Stack(x, 0)$   \Comment{Stack the blocks along $0$-th dimension.}
        \State $i \leftarrow i+1$
    \EndWhile  
\EndProcedure

\end{algorithmic}
\label{alg:data_decomposition}
\end{algorithm}

\begin{algorithm}[h]
\caption{Window patching}
\begin{algorithmic}[1]

\Procedure{Window-patching}{$x,b, B, d$}       \Comment{domain tensor, batch size, block numbers, dimension number}
    \State $i \leftarrow 0$
    \While{$i \not= d$}  
        \If{$d-i-2 < 0$}
            \State $V \leftarrow b$
        \Else
            \State $ V \leftarrow b\times \prod_{j=0}^{d-i-2} B_i $
        \EndIf
        \State $x \leftarrow Split(x,V,0)$  \Comment{Split the x along $0$-th dimension into $V$ blocks.}
        \State $x \leftarrow Stack(x, d-i)$   \Comment{Stack the blocks along $(d-i)$-th dimension.}
        \State $i \leftarrow i+1$
    \EndWhile  
\EndProcedure

\end{algorithmic}
\label{alg:window_patching}
\end{algorithm}

\begin{algorithm}[h]
\caption{Prediction integration}
\begin{algorithmic}[1]

\Procedure{Prediction-integration}{$x,NN,w,b,B,N,P,d$}       \Comment{domain tensor, neural network, window size, batch size, block number, domain size, window points, dimension number}
    \State $x \leftarrow \text{Expand-Domain}(x,w,N)$    \Comment{Expand the domain by padding zeros}

    \For{$p \in P$}    \Comment{Loop over all the points in a window}
        \State $x_p \leftarrow x[p:p+wB]$.   \Comment{Select the part of $x$ that starts from $p$ with size $wB$}   
        \State $x_p \leftarrow \text{Chunk-Domain}(x_p, B, d)$   \Comment{Decompose the data}
        \State $y_p \leftarrow NN(x_p)$    \Comment{Make the prediction over the decomposed data}
        \State $y_p \leftarrow \text{Window-Patch}(y_p,b, B, d)$   \Comment{Recover the domain to the original shape}
        \State $\{y\} \leftarrow y_p[w/2]$    \Comment{Store the values at the window centers of $y_p$}
    \EndFor

\EndProcedure

\end{algorithmic}
\label{alg:prediction_integration}
\end{algorithm}

\subsection{Window size determination}\label{sec:window_size}

 We define $L_c$ for hyperbolic PDEs in a similar way to the Courant–Friedrichs–Lewy (CFL) condition \cite{de2013courant} in computational fluid dynamics which is a necessary condition for convergence while solving hyperbolic PDEs by guaranteeing the subdomain contains enough information about the \textit{flow of information}. CFL condition specifies an upper bound for the time step $\Delta t$ with Courant number $C$ as
\begin{equation}
    C = \frac{c\delta t}{\delta x} \leq C_{max},
\end{equation}
where $c$ is the transport velocity, $\delta x$ is the element length of the discretization, $C_{max}$ is decided by experience which could range between 0.1 to 100 according to different solvers. 

Similarly, we can define a lower bound for the distance the physical information travels with a fixed $\Delta t$, which is called the characteristic length $L_c$. $L_c$ can be determined according to the physical coefficients relevant to PDEs. For example, for mass transport PDEs, we have $L_c = c \Delta$ where $c$ is the mass transport speed whose base unit is $[L/T]$. For heat transfer PDEs, we have $L_c = \sqrt{\alpha \Delta}$ where $\alpha$ is the thermal diffusivity whose base unit is $[L^2/T]$. For Burger's equation, there are two terms related to momentum transport, the convection and the diffusion terms, which have their own characteristic lengths respectively. For the convection term, we have $L_c^{conv}=u\Delta t$, where $u$ is the fluid velocity, which is similar to mass transport speed. For the diffusion term, we have $L_c^{diff}=\sqrt{\nu \Delta t}$, where $\nu$ is the diffusion coefficient, which has the same base unit as temperature diffusivity $\alpha$. We determine the characteristic length for Burger's equation as $L_c = \max(L_c^{conv},L_c^{diff})$. The element length $\delta x$ of the discretization in our numerical simulation data is in the same scale of $L_c$. 

From our experience, when the window size is in the scale of $\backsim 10 L_c$, DDELD usually has the best effects in improving the model's prediction accuracy. Given a specific dataset, we could not specify the optimal window size without a hyper-parameter tuning process because of two reasons. Firstly, the physical coefficients used to determine the characteristic lengths cannot precisely reflect the information traveling speed for all the points in the domain as the speed is affected by many other factors. For example, for mass transport, besides $c$, the transport speed is also affected by the gradients of mass concentration. Secondly, the initial and boundary conditions of the data would also affect the features of PDEs. Like all the data-driven methods, the hyper-parameters of the models could only be tuned through experience for data with various features. 

What we find through experiments, however, indicates an alternate approach to finding a desired window size by analyzing the frequency distribution of the solution if we have some knowledge of it \textit{a priori}. We find that data-driven methods for learning mappings between PDE operators are often more sensitive to the structure of the initial condition (or generally the input function to the model) than the coefficients in the PDE itself, see Appendix for a detailed explanation. This highlights the possibility of a frequency-based analysis for the choice of window size for PDEs. We give the following theorem as a guideline for the choice of window sizes for generic physical data with a clear frequency bias in Theorem~\ref{theorem:frequency_based_domain} and appended the proof in Appendix~\ref{appendix:proof}. 

\begin{theorem}\label{theorem:frequency_based_domain}
    Define $u\in L_1(\mathbb{R})$ and let $\hat{u}$ be the Fourier transform of $u$ with a clear frequency bandwith of $\text{supp}(\hat{u})\subseteq [-B, B]$. If a set of observations of discretized values of $u$ per \textit{unit length} $\{u(x_i)|i=1, 2, 3, \cdots, N\}$ are available, and we obtain a local dependency region based on Equation~\ref{equ:segmentation}, then the \textit{minimum} number of points required for the local dependency region to retain frequency character would be given by: $L_c=\lceil{\frac{N+1}{2B}}\rceil$.
\end{theorem}
The experimental results regarding the influence of window size and information frequency over the model's prediction accuracy is shown in Section \ref{sec:freq_results}.

\section{Data Generation}\label{appendix:data_generation}
In this section, we describe the generating process of the datasets used to evaluate the effects of DDELD, including 2D mass transport equation, 2D Burger's equation, 2D isotropic turbulence in fluid dynamics, and 3D temperature transferring data.

\subsection{Mass transport equation}
\label{sec:transport}
The transport of mass can be seen as one of the most fundamental PDEs with variation in both time and space. It also enjoys the benefit of having a fully closed mathematical solution, and that the problem can be carefully constructed to show the solution field of different character frequencies. We test the DDELD's representation ability on solution functions of different frequencies to understand the decomposition method's performance on questions including how small the decomposition can get, and how wide the frequency range the decomposition can capture before the model starts to lose accuracy due to domain cut-offs in Section \ref{sec:results}.
A typical mass transport equation can be expressed as Equation~\ref{equ:mass_trans}:
\begin{equation}
    \frac{\partial u}{\partial t}=-c\cdot\nabla\mathbf{u},
    \label{equ:mass_trans}
\end{equation}
and the exact mathematical equation can be written as Equation 
\begin{equation}
    \mathbf{u(t)}=\mathbf{u}_0(x-ct), 
\end{equation}
for any initial condition $\mathbf{u}_0$, transport speed $c$ and given temporal stamp $t$. We can thus construct the frequency of our solution by determining the frequency of the initial condition.
Variants of the solution frequency are shown in Figure \ref{fig:data_transport}.
\begin{figure*}[h]
    \centering
    \includegraphics[width=0.9\textwidth]{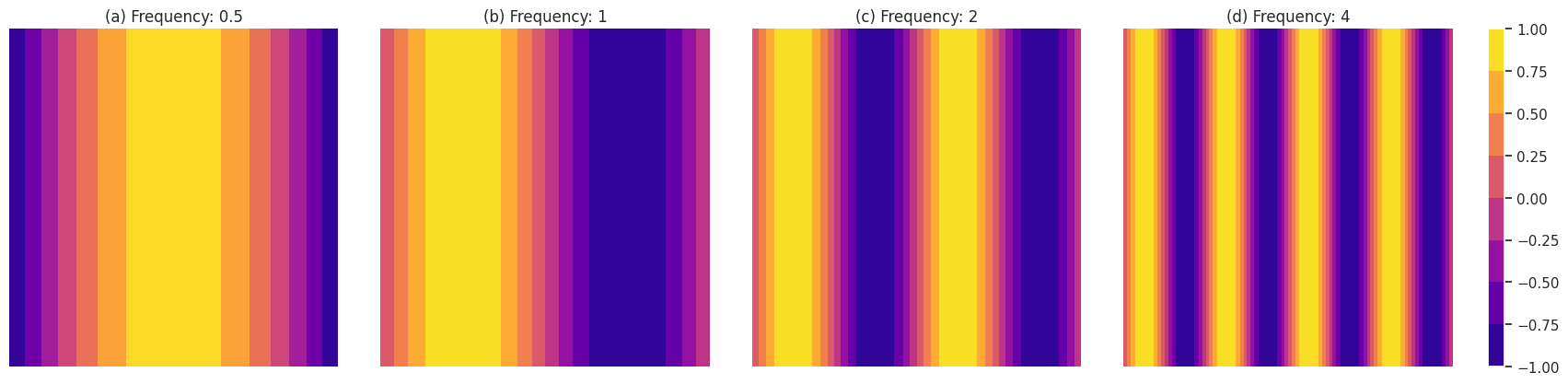}
    \caption{Solution of the mass transport equation of a $\sin$ wave in different frequencies. All displayed physical properties are normalized and dimensionless. (a) $f=0.5$; (b) $f=1.0$, (c) $f=2.0$, (d) $f=4.0$}
    \label{fig:data_transport}
\end{figure*}

\subsection{Burgers' equation}
\label{sec:burgers}
The Burgers' equation is also one of the most representative PDEs representing a convection-diffusion scheme. The solution of such an equation displays certain interesting physical phenomena including temporal wave propagation, shock wave formulation, and viscous-related energy dissipation. Approximating these complex dynamics is a challenging task for a machine learning model with no \textit{a priori} information about the underlying physics, and therefore makes it a good testing case for validating model performance. 

We implement the viscous version of Burgers' equation as described by Equation  \ref{eq:burgers}:
\begin{equation}
    \frac{\partial \mathbf{u}}{\partial t}+(\mathbf{u}\cdot\nabla)\mathbf{u}=\nu\nabla^2 \mathbf{u},~\mathbf{x}\in D
    \label{eq:burgers}
\end{equation}
where $\mathbf{u}$ denotes the velocity of the fluid, $\mathbf{x}$ and $t$ are spatial and temporal coordinates respectively, and $\nu$ is the viscosity of the fluid.

We solve Equation \ref{eq:burgers} with $\nu=0.01 Pa\cdot s$, and a time step of $0.1s$ for a total of 10 seconds with randomly initialized velocity Gaussian distribution as the initial condition. The simulations of the Burgers' equation under different initial conditions are performed in FEniCS on a 2D mesh of $80\times 80$ elements per unit. Four Burgers' equation solutions computed in four different initial velocity distributions are shown in Figure \ref{fig:data_burgers}.
\begin{figure*}[h]
    \centering
    \includegraphics[width=0.9\textwidth]{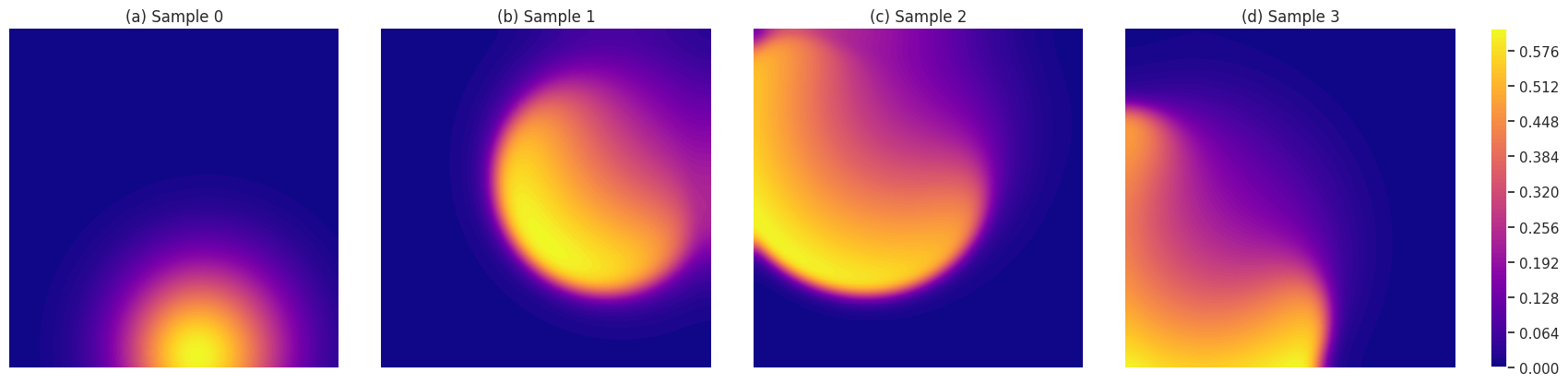}
    \caption{Solution of the Burgers' equation in 2D with random initialization. All displayed physical properties are normalized and dimensionless.}
    \label{fig:data_burgers}
\end{figure*}

\subsection{Isotropic turbulence}
We performed testing and validation of the DDELD on simulation results with direct numerical simulation (DNS) on an isotropic turbulence scenario. The incompressible Navier-Stokes equation can be written as Equation~\ref{eq:ns}:
\begin{equation}
    \frac{\partial \mathbf{u}}{\partial t}+(\mathbf{u}\cdot\nabla)\mathbf{u}=\nu\nabla^2 \mathbf{u}-\frac{1}{\rho}\Delta p+\mathbf{f},~\mathbf{x}\in D
    \label{eq:ns}
\end{equation}
where $\nu$ denotes the fluid viscosity, $\rho$ is the fluid density and $\mathbf{f}$ denotes the body force. We obtained DNS data from the Johns Hopkins Turbulence Database~\cite{li2008public} (JHTDB) with a total of 5,028 sequential time steps solved with a pseudo-spectral solver. We present samples of the turbulent data in Figure~\ref{fig:data_jhtdb}
\begin{figure*}[h]
    \centering
    \includegraphics[width=0.9\textwidth]{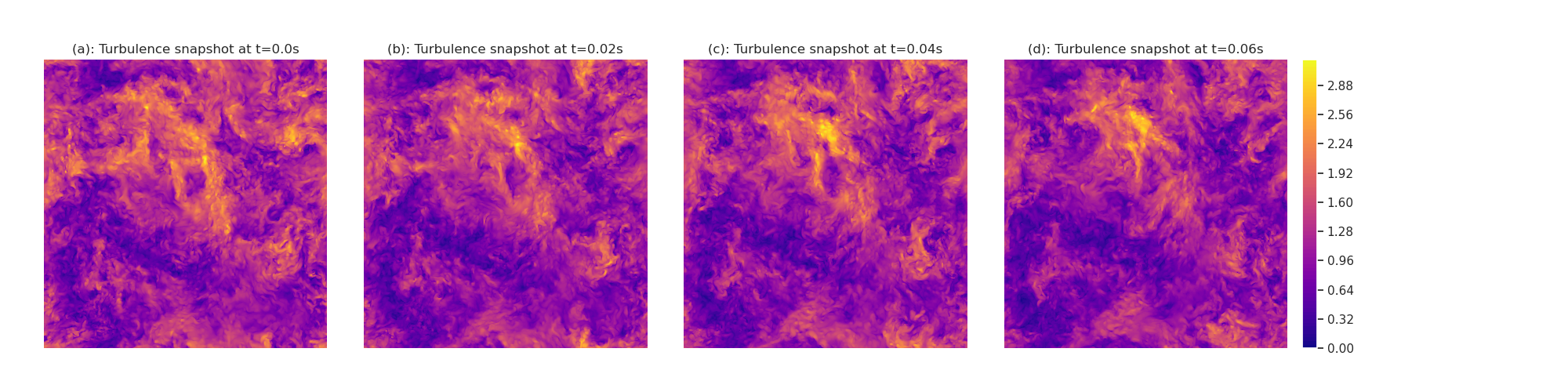}
    \caption{Solution of the isotropic turbulence at t=0s, t=0.02s, t=0.04s and t=0.06s respectively.}
    \label{fig:data_jhtdb}
\end{figure*}

\subsection{AM temperature numerical simulation}\label{appendix:am_temperature}
Thermal simulations for metal additive manufacturing (AM) are important in multiple stages of product development that involve AM processes, including part design, process planning, process monitoring, and process control \cite{sun2022thermodynamics,mozaffar2023differentiable,chen2023capturing, chen2024accelerating}. Because of the geometric complexity of the parts, AM thermal simulation is the typical scenario where the traditional numerical methods are too time-consuming while the data-driven ML models are difficult to generalize to the situations not included in the training data. So, it is critical to increase the geometric generalizability of data-driven models with limited data. Here, we generated an AM temperature dataset including the temperature histories of 10 parts in different geometries to evluate the effects of DDELD in improving the geometric generalizability of data-driven models. The 10 geometries are randomly generated by SkexGen \cite{xu2022skexgen}, which contains various common mechanical features, such as holes, ribs, and pillars. Figure \ref{fig:am_temp_data} shows the examples of the temperature data.
\begin{figure}[h]
    \centering
    \includegraphics[width=0.4\linewidth]{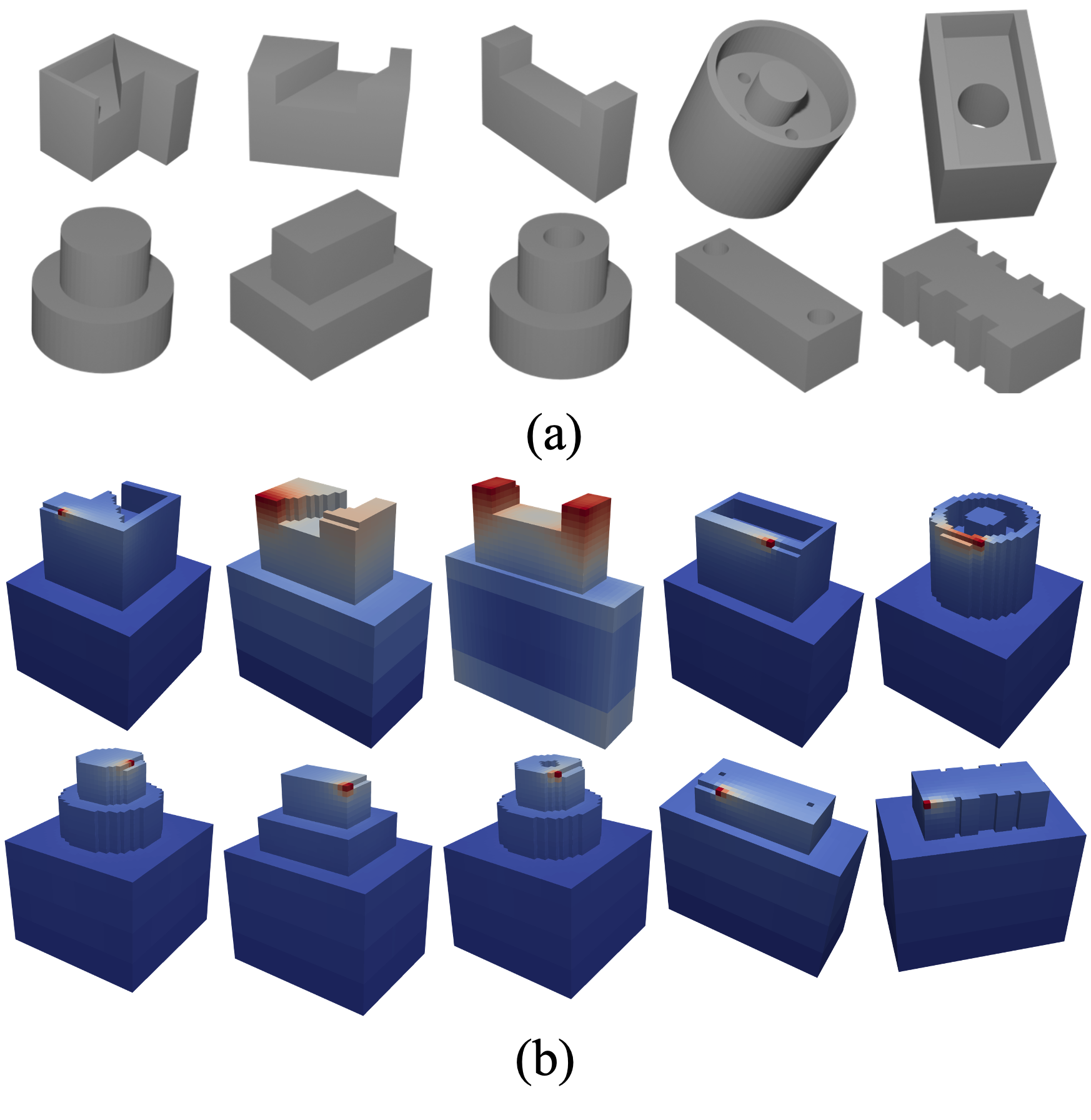}
    \caption{AM temperature prediction dataset. (a) The 10 parts with various geometries. (b) The temperature histories of the AM process that built the 10 parts are generated.}
    \label{fig:am_temp_data}
\end{figure}

The heat transfer PDEs for AM process is formulated as follows.
Let $\Omega$ denote a domain, and $\partial \Omega$ its boundary. $\partial \Omega_{\mathrm{H}}$ represents the part at which heat is transferred to the surroundings with constant temperature $T_{\infty}$, and $\partial \Omega_{\mathrm{D}}$ the part at which the temperature is fixed at $T_{\mathrm{D}}$. The temperature evolution within $\Omega$ is governed by the heat transfer PDEs:
\begin{equation}\label{equ:thermal_pdes}
\begin{aligned}
\rho c_p \dot{T} &= \nabla \cdot (k_p \nabla T), &&\forall \mathbf{x} \in \partial \Omega  \\
-\mathbf{n} \cdot k_p \nabla T &= h_{\mathrm{c}}(T-T_{\infty}), &&\forall \mathbf{x} \in \partial \Omega_{\mathrm{H}} \\
T &= T_{\mathrm{D}}, &&\forall \mathbf{x} \in \partial \Omega_{\mathrm{D}}.
\end{aligned}
\end{equation}
Here, $\rho$, $c_{\mathrm{p}}$, and $k_{\mathrm{p}}$ are the temperature-dependent density, specific heat capacity, and conductivity of the material, respectively. The vector $\mathbf{n}$ is the unit outward normal of the boundary at coordinate $\mathbf{x}$. 

We utilized the thermal simulation algorithm developed in \cite{nijhuis2021efficient} to solve the partial differential equations described in Equation \ref{equ:thermal_pdes} for the wired-based DED process. This algorithm uses the discontinuous Galerkin FEM to spatially discretize the problem and the explicit forward Euler time-stepping scheme to advance the solution in time. The algorithm activates elements based on the predefined toolpath. Newly deposited elements are initialized at elevated temperatures, after which they are allowed to cool according to Equation \ref{equ:thermal_pdes}. The temperature of the substrate’s bottom face is kept fixed at $T_\infty =25^\circ C$.  On all other faces, convection, and radiation to the surrounding air at $T_\infty$ is modeled. We set the tool moving speed to $5 mm/s$. All geometric models were discretized with a resolution of $20\times20\times20$, with an element size of $2 mm$. Our simulations utilized S355 structural steel as the material, with material properties as given in \cite{nijhuis2021efficient}.

\section{Numerical Experiments}\label{sec:experiments}
DDELD is implemented over CNN, FNO, Multiwavelet-based Operator (MWO) \cite{gupta2021multiwavelet}, DeepONet \cite{lu2019deeponet}, and Dil-ResNet \cite{stachenfeld2021learned} in our experiments. In the case of DeepONet, two variations of encoding neural networks are employed for its branch net: fully-connected (FC) networks, and CNNs. This yields a total of six baseline models: CNN, FNO, MWO, DeepONet-FC, DeepONet-CNN, and Dil-ResNet. FC and CNN are two classical neural network architectures. FC lacks local-dependency, whereas CNN exhibits limited local-dependency within a single layer. FNO and DeepONet stand as state-of-the-art neural operators for PDE problems, yet they do not inherently possess local-dependency. Dil-ResNet is the state-of-art model for turbulence simulation. 

The FC, FNO and CNN models applied in our experiments each have 4 layers with 20 hidden dimensions. The Adam optimizer \cite{kingma2014adam} is employed with a learning rate of $0.0001$. Normalized $L_2$ is used as training and testing loss and the $R^2$ score is applied for validation metric. A 50/50 train/test split is applied to push the generalizabilities of the neural operators on the edge. Code is publicly available on \footnote{\href{https://drive.google.com/drive/folders/1ZpYUl5kaymZEH9H23y0pgMRQNl6fVlV-?usp=drive_link}{Google drive}}. The processing speed analysis of the algorithm is presented in Appendix \ref{sec:time_cost}.

Normalized $L_2$ and $R^2$ are used as evaluation metrics for the prediction at each time step, which are defined as follows. 
\begin{equation}
\begin{aligned}
    &L_2 = \sum_{i=1}^n \frac{\sqrt{(u_{\mathrm{pred}_i} - u_i)^2}}{|u_i|},
    &R^2 =  1 -  \frac{\sum_{i=1}^n (u_{\mathrm{pred}_i} - u_i)^2}{\sum_{i=1}^n (u_{\mathrm{mean}} - u_i)^2},\\
\end{aligned}
\end{equation}
where $u_{\mathrm{pred}_i}$ and $u_i$ represent the predicted and ground truth temperature of an individual position, respectively, $n$ represents the element number in a discretization of domain, and $u_{\mathrm{mean}}$ represents the average values in over the domain. $L_2$ can reflect the mean accuracy of the ML model, and $R^2$ measures the proportion of the variance in the ground truth that is explained by prediction. Note that the average of the metrics over the test dataset are calculated in the following discussion.

\section{Results and Discussion}
\label{sec:results}

\subsection{DDELD for accelerating training convergence}
\begin{figure}
    \centering
    \includegraphics[width=\linewidth]{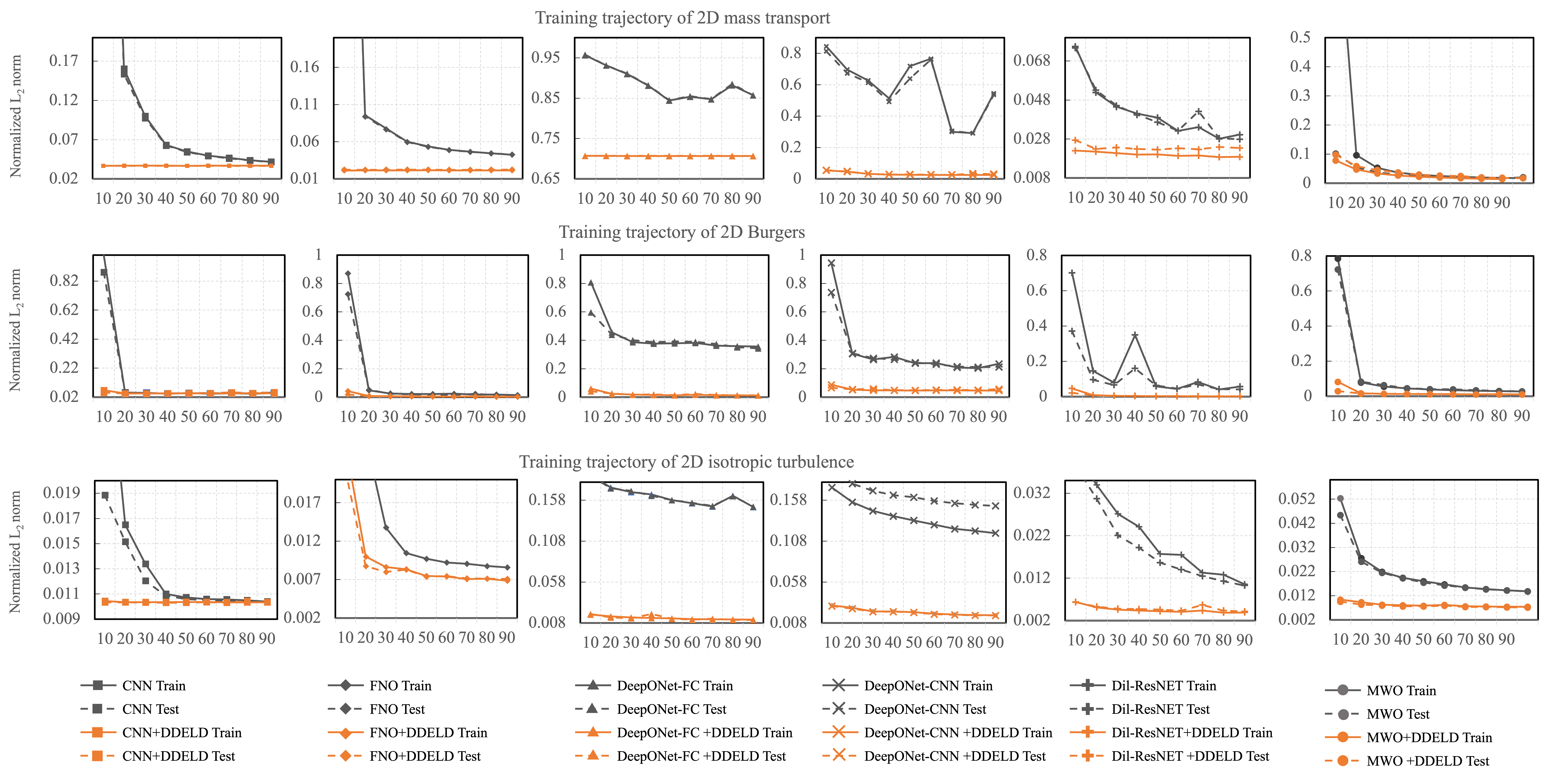}
    \caption{DDELD accelerates training convergences of the 5 models over the 3 datasets. The model errors are evaluated by normalized $L_2$ errors.}
    \label{fig:training_trajectory}
\end{figure}

\newcolumntype{R}{>{\raggedright \arraybackslash} X}
\newcolumntype{S}{>{\centering \arraybackslash} X}
\newcolumntype{T}{>{\raggedleft \arraybackslash} X}
\begin{table}[]
\tiny
    \centering
    \begin{tabularx}{\linewidth} {>{\setlength\hsize{.1\hsize}}R >{\setlength\hsize{.0025\hsize}}S >{\setlength\hsize{.0879\hsize}}S >{\setlength\hsize{.0879\hsize}}S >{\setlength\hsize{.0879\hsize}}S >{\setlength\hsize{.0879\hsize}}S >{\setlength\hsize{.0879\hsize}}S >{\setlength\hsize{.0879\hsize}}S >{\setlength\hsize{.0879\hsize}}S > 
    {\setlength\hsize{.0879\hsize}}S >{\setlength\hsize{.0879\hsize}}S >
    {\setlength\hsize{.0879\hsize}}S}  
    \toprule
         \multirow{2}{*}{} & \multirow{2}{*}{} & \multicolumn{2}{c}{CNN} & \multicolumn{2}{c}{FNO} & \multicolumn{2}{c}{MWO} & \multicolumn{2}{c}{DeepONet-CNN} & \multicolumn{2}{c}{Dil-ResNet}  \\
         \cmidrule(l){3-4} \cmidrule(l){5-6} \cmidrule(l){7-8} \cmidrule(l){9-10}\cmidrule(l){11-12}\\
         \multicolumn{2}{c}{DDELD}& No & Yes & No & Yes & No & Yes & No & Yes & No & Yes\\
         \hline
         Mass & $L_2$ & 0.0414 & \colorbox{lightgray}{0.0366} & 0.0427& \colorbox{lightgray}{0.0221} & 0.0200 & \colorbox{lightgray}{0.0172} & 0.5365 & \colorbox{lightgray}{0.0326} & 0.0278 & \colorbox{lightgray}{0.0229} \\
         transport & $R^2$ & 0.9975 & \colorbox{lightgray}{0.9978} & 0.9978 & \colorbox{lightgray}{0.9994} & 0.9997 & \colorbox{lightgray}{0.9998} & 0.6794 & \colorbox{lightgray}{0.9989} & 0.9990 & \colorbox{lightgray}{0.9992} \\
         \cmidrule(l){1-2}
         \multirow{2}{*}{Burger's} & $L_2$ & 0.0511& \colorbox{lightgray}{0.0456} & 0.0137& \colorbox{lightgray}{0.0061} & 0.0263 & \colorbox{lightgray}{0.0111}& 0.2117 & \colorbox{lightgray}{0.0552} & 0.0413 & \colorbox{lightgray}{0.0059} \\
         & $R^2$& 0.9936 & \colorbox{lightgray}{0.9943} & 0.9997 & \colorbox{lightgray}{0.9999} & 0.9988 & \colorbox{lightgray}{0.9997} & 0.8988 & \colorbox{lightgray}{0.9941} & 0.9974 & \colorbox{lightgray}{0.9999} \\
         \cmidrule(l){1-2}
          Isotropic&$L_2$& 0.0104& \colorbox{lightgray}{0.0103}& 0.0086& \colorbox{lightgray}{0.0079}& 0.0137& \colorbox{lightgray}{0.0071}& 0.1510 & \colorbox{lightgray}{0.0217} & 0.0102 & \colorbox{lightgray}{0.0047} \\
         turbulence& $R^2$& 0.9981 & \colorbox{lightgray}{0.9982} & 0.9989 & \colorbox{lightgray}{0.9993} & 0.9968 & \colorbox{lightgray}{0.9992} & 0.8123 & \colorbox{lightgray}{0.9959} & 0.9973 & \colorbox{lightgray}{0.9997} \\
         \hline
         \multicolumn{2}{c}{Inf. time (ms)} & 5.097  & 12.099 & 5.485 & 16.684 & 7.709 & 14.027 & 7.353 &19.872 & 10.530 & 18.914 \\
         \multicolumn{2}{c}{\# params.} & \multicolumn{2}{c}{3,581}& \multicolumn{2}{c}{104,241}& \multicolumn{2}{c}{10,841}& \multicolumn{2}{c}{590,641}& \multicolumn{2}{c}{103,421} \\
         \bottomrule
    \end{tabularx}
    \caption{Evaluation results of the 5 models over the 3 datasets with and without DDELD. The DDELD window size for the 3 datasets are selected as 11,9 and 11, respectively.}
    \begin{tablenotes}
      \scriptsize
      \item \textit{Note 1}. In our experiments, we used a 50/50 training/test split, while the benchmark models in the literature typically use a 90/10 split. We selected a 50/50 split to push the generalizability of the benchmark models to the edge when a training process that can efficiently capture local physical evolution patterns is most needed.\\
    \end{tablenotes}
    \begin{tablenotes}
      \scriptsize
      \item \textit{Note 2}. To compare the models trained by similar wall-clock time, the reported results of the models with DDELD are trained with 30 epoches, while the results of the model without DDELD are trained with 90 epoches.\\
    \end{tablenotes}
    \label{tab:evaluation_results}
\end{table}

DDELD can help accelerate the convergence of the ML models for time-dependent PDEs with local-dependency by limiting the scope of input data. Figure \ref{fig:training_trajectory} shows the training and test $L_2$ history during the training process for the ML models with and without DDELD over the mass transport,  burger's, and  isotropic turbulence. 
As we can see, over all the cases, the training processes with DDELD can reduce the test errors faster than the processes without DDELD. These results confirm our assumption that the coupling between the expressiveness and local dependency of deep learning architecture may expand the scope of input data beyond the travel range of the physics information, which can result in sluggish convergence. On the other hand, strictly limiting the scope of input data can speed up the training process. 

The effects of the DDELD may differ depending on the type of ML model. From he final test errors shown in Table \ref{tab:evaluation_results}, we can see that FNO, DeepONet-FC, DeepONet-CNN and Dil-ResNet gain a much larger improvement in the convergence rate than CNNs do. Such a difference might originate from their local dependency. The linear operator of FNOs is the convolution in Fourier space, which involves the integral over the whole domain. DeepONet-FC, DEEpONet-CNN and Dil-ResNet encodes the global information of the whole domain. So, they are not local-dependent. DDELD furnishes them with strict local-dependency while does not changing their structures. So, the improvements solely come from the limiting of the data scope. On the other hand, the linear operator of CNNs is a kind of local convolution whose scope is prescribed by its kernel size. While the deep learning architecture can weaken its local dependency, CNNs are still more local-dependent than the non-local-dependent operators. These results supports our assumption that more data does not always beget better results. Especially, the information outside of the local-dependent region is irrelevant to the local physical evolution. This extra information is essentially noises, which hinders the ML model from capturing the real physical patterns in the data. Therefore, limiting the scope of input data can effectively filter out the noises and help the ML models capture the real physical patterns contained in the data.

The improvements in prediction accuracy achieved by DDELD are directly illustrated in Figure \ref{fig:ns_prediction} (a). This figure shows examples of 2D isotropic turbulence solutions at $0.1 s$, $0.2 s$, and $0.3 s$ from the test dataset, along with their corresponding predictions made by FNO and FNO with DDELD. The window size is $11 \delta x$. As demonstrated, DDELD significantly reduces FNO's prediction error. Similar plots for the 2D mass transport and 2D burger's equation can be found in Figures \ref{fig:mass_prediction} and \ref{fig:burges_prediction}.

The method discussed in Section \ref{sec:window_size} estimates the range of the optimal window size based on the physical characteristics and time step, which is on the order of $10 \delta x$. Theoretically, there exists a window size that minimizes the approximation error of a neural operator for specific PDEs. However, in addition to approximation error, ML models also have optimization error and generalization error \cite{lu2019deeponet}, which depend on training strategies and datasets. Therefore, the window size should be fine-tuned as a hyper-parameter in practice, as shown in Figure \ref{fig:ns_prediction} (b).
\begin{figure}
    \centering
    \includegraphics[width=\linewidth]{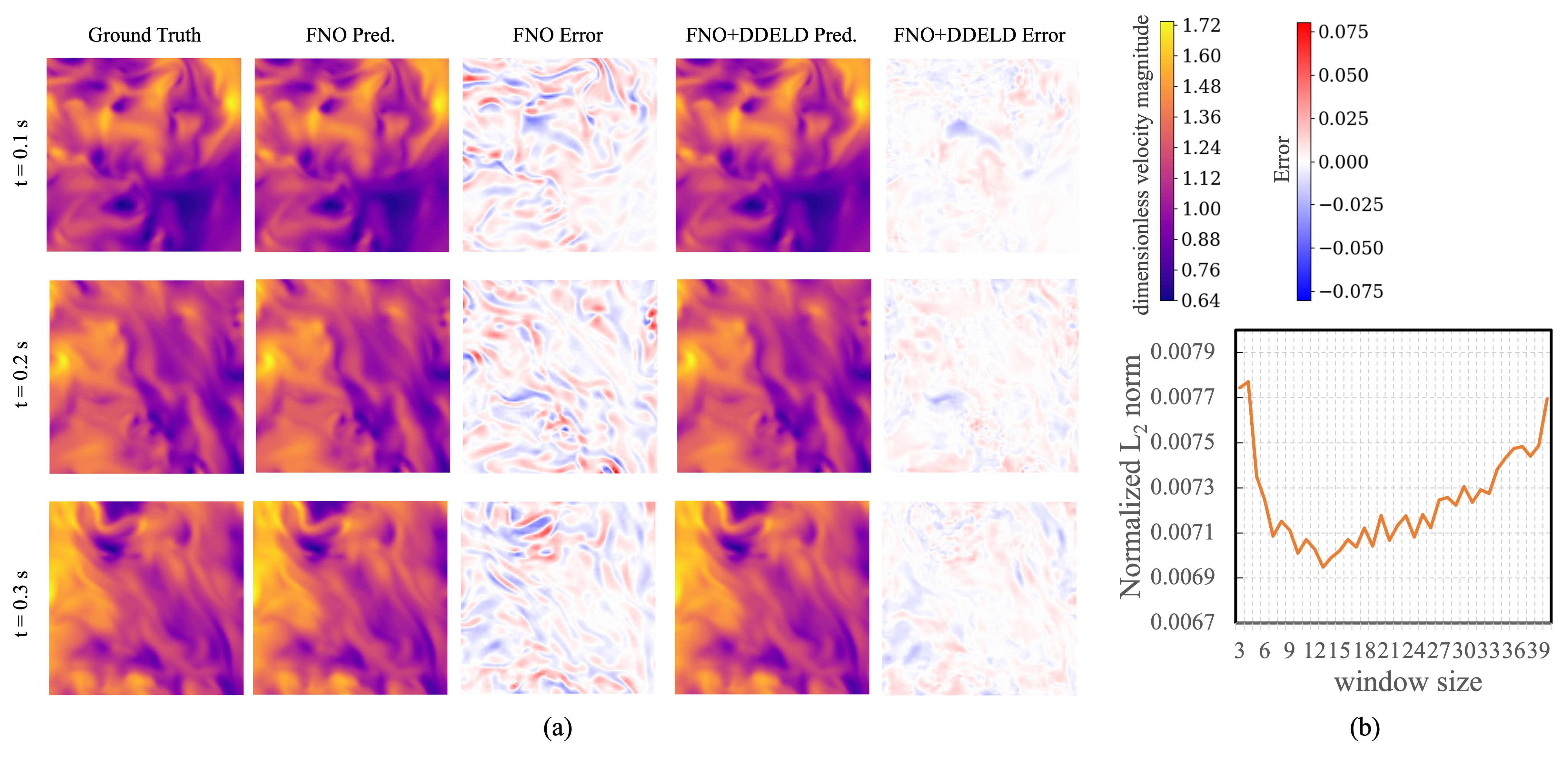}
    \caption{(a) Examples of isotropic turbulence predictions made by FNO and FNO with DDELD. All displayed physical properties are normalized and dimensionless. (b) The test errors measured in normalized $L_2$ norm of FNO with DDELD in various window sizes over isotropic turbulence data.  }
    \label{fig:ns_prediction}
\end{figure}

\begin{figure}
    \centering
    \includegraphics[width=0.8\linewidth]{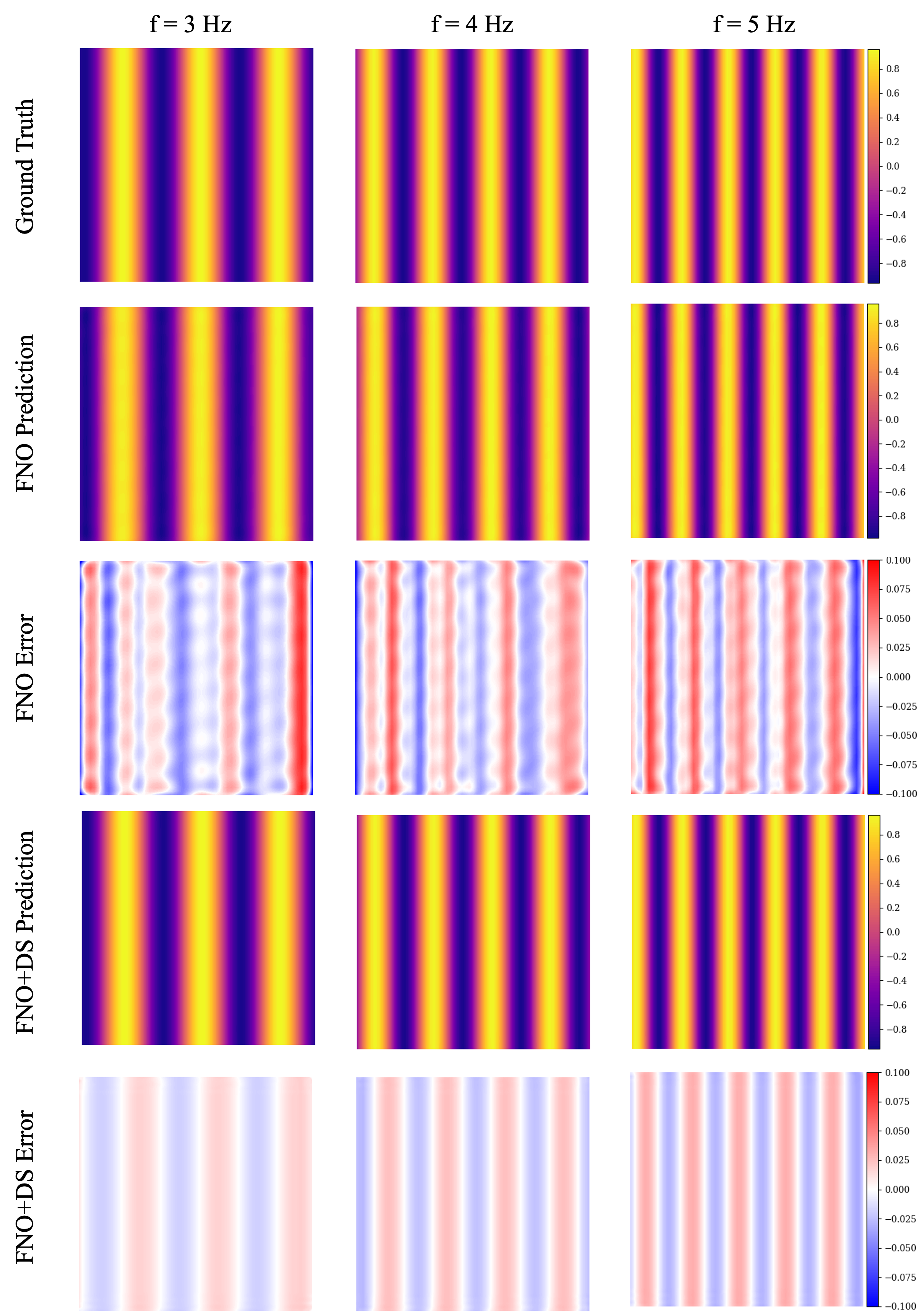}
    \caption{Examples of mass transport predictions in $3 Hz$, $4 Hz$, and $5 Hz$ made by FNO and FNO with DDELD. The window size is $11 \delta x$.  All displayed physical properties are normalized and dimensionless. }
    \label{fig:mass_prediction}
\end{figure}

\begin{figure}
    \centering
    \includegraphics[width=0.8\linewidth]{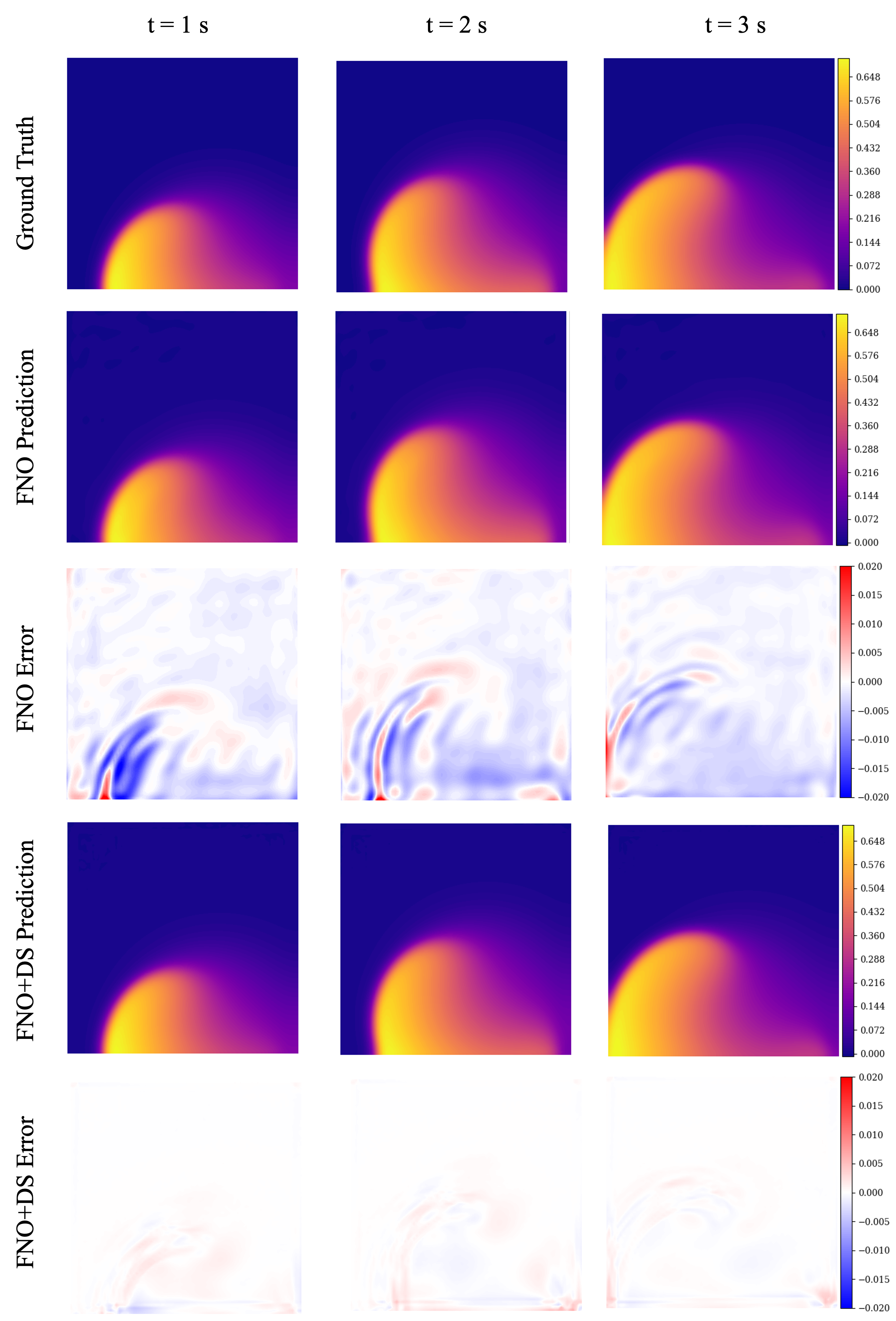}
    \caption{Examples of Burgers' equation predictions in $1 s$, $2 s$, and $3 s$ made by FNO and FNO with DDELD. The window size is $9 \delta x$. All displayed physical properties are normalized and dimensionless. }
    \label{fig:burges_prediction}
\end{figure}

\subsection{DDELD for improving geometric generalizability}
We test the capability of DDELD method in improving the geometric generalizability of the ML models over the 3D AM temperature data including the temperature histories of 10 parts as shown in Figure \ref{fig:am_temp_data}. A 10-fold leave-one-outside cross-validation (LOOCV) is performed to evaluate the $R^2$ accuracy of the ML models over the geometry not included in the training data. At each round of LOOCV, the data from 9 parts are assembled as training and test data with a 9:1 ratio, and the data from the rest of one part is used for validation. The window size is $7 \delta x$. Examples of the prediction results can be found in Figure \ref{fig:am_prdiction_example}.  Figure \ref{fig:am_prediction} shows the examples of AM heat transfer equation solutions in the test dataset, and their predictions made by FNO and FNO with DDELD. As we can see, DDELD significantly reduces the errors of the temperature prediction whose part geometry is not included in the training process.

\begin{figure}
    \centering
    \includegraphics[width=0.8\linewidth]{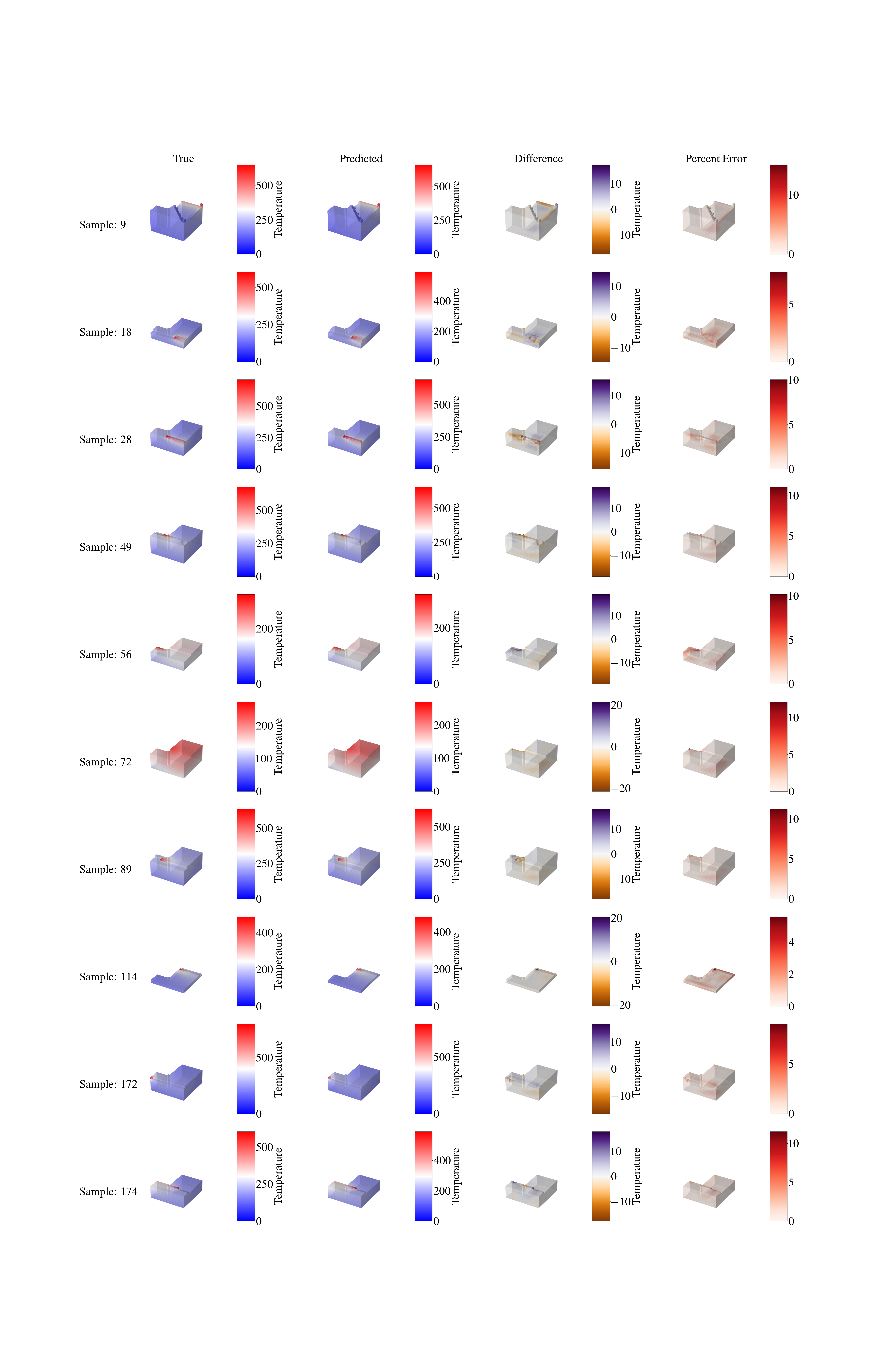}
    \caption{AM temperature prediction made by FNOs with DDELD. The samples are randomly selected from the validation data of part 1.}
    \label{fig:am_prdiction_example}
\end{figure}

\begin{figure}[h]
    \centering
    \includegraphics[width=0.8\linewidth]{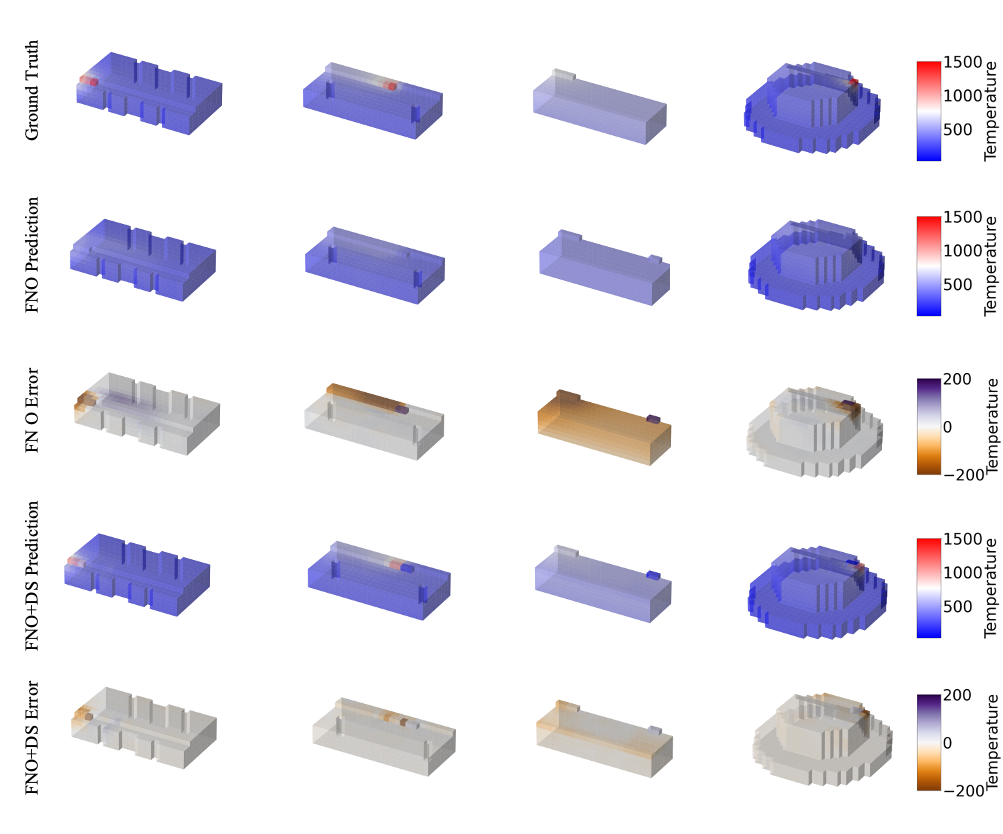}
    \caption{Examples of AM heat transfer equation predictions made by FNO and FNO with DDELD. The temperature unit is $^{\circ}C$. }
    \label{fig:am_prediction}
\end{figure}

\begin{figure}[h]
    \centering
    \includegraphics[width=0.6\linewidth]{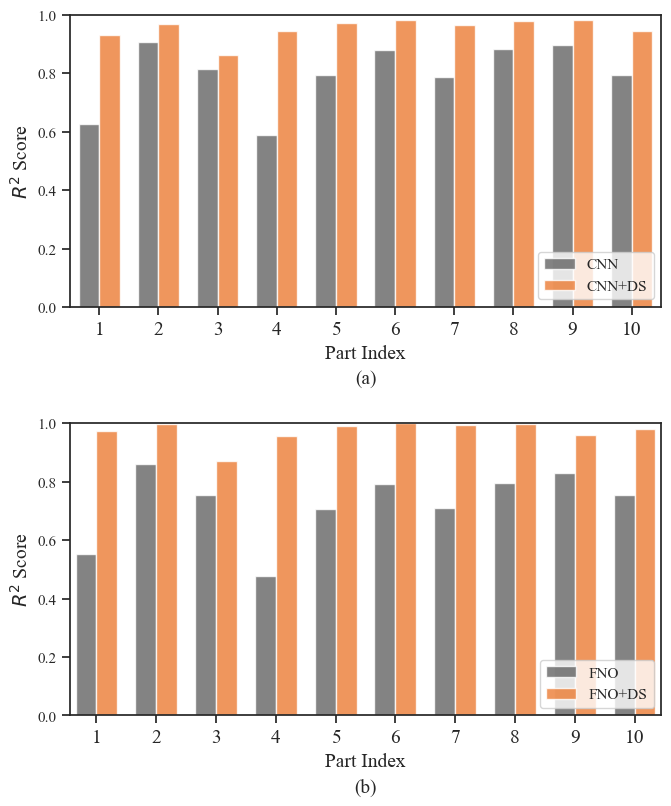}
    \caption{DDELD improves the geometric generalizability of the ML models for temperature prediction during AM processes. There are 10 rounds of leave-one-out cross-validation. In each round, the data of one geometric part is taken out and the ML models are trained on the data of the other 9 parts. Then, the ML models are validated over the data of the part not included in the training. }
    \label{fig:am_thermal_r2}
\end{figure}

DDELD method can improve the geometric generalizability of the ML models, as shown in Table \ref{tab:full_data} where the test $L_2$ and $R^2$ over the models with and without DDELD are listed. Figure \ref{fig:am_thermal_r2} plots the validation $R^2$ of the ML models over the 10-fold LOOCV of the AM temperature prediction dataset. The part index in the horizontal axis indicates the 10 parts with different geometries that are not included in the training and test of each validation round. So, the validation $R^2$ reveals the geometric generalizability of the models. As we can see, the DDELD enhances the accuracy of all the validation data for both of the models. On average, CNNs are improved by 21.7 \%, and FNOs are improved by 38.5\%.

\newcolumntype{R}{>{\raggedright \arraybackslash} X}
\newcolumntype{S}{>{\centering \arraybackslash} X}
\newcolumntype{T}{>{\raggedleft \arraybackslash} X}
\begin{sidewaystable}
\footnotesize
    \renewcommand{\arraystretch}{1.2}
    \centering
    \caption{The test $L_2$ and $R^2$ of the models with and without DDELD over AM heat transfer datasets.}  
    \medskip 
    \begin{tabularx}{\linewidth} {>{\setlength\hsize{.0025\hsize}}R >{\setlength\hsize{.1\hsize}}S  >{\setlength\hsize{.0625\hsize}}S >{\setlength\hsize{.0625\hsize}}S >{\setlength\hsize{.0625\hsize}}S > 
    {\setlength\hsize{.0625\hsize}}S >{\setlength\hsize{.0625\hsize}}S >
    {\setlength\hsize{.0625\hsize}}S >{\setlength\hsize{.0625\hsize}}S >
    {\setlength\hsize{.0625\hsize}}S > {\setlength\hsize{.0625\hsize}}S >  {\setlength\hsize{.0625\hsize}}S}  
   \toprule
    & Models &  \multicolumn{10}{c}{10-fold LOOCV AM heat transfer} \\
        &  & Part 1 & Part 2 &Part 3 &Part 4 &Part 5 &Part 6 &Part 7 &Part 8 &Part 9 &Part 10 \\
        \cmidrule(l){3-12}\\
        \multirow{4}{*}{$L_2$} & FNO & 3.435 & 3.486 & 3.227 & 3.292 & 3.506 & 3.678 & 3.536 & 3.684 & 3.537 & 3.566 \\
        & FNO+DDELD & 0.2452& 0.2578&0.1387& 0.2094& 0.19308& 0.2689& 0.2289& 0.3151& 0.1889& 0.1679\\
        & CNN & 2.617 & 2.847& 2.623& 2.474& 2.296&2.396 & 3.032& 2.552& 2.272& 2.991\\
        & CNN+DDELD & 2.748& 2.286& 2.214& 2.434& 2.590& 2.597& 3.032& 2.755& 2.162& 2.205\\
        \multirow{4}{*}{$R^2$} & FNO & 0.5506 & 0.8594 & 0.7557& 0.4753& 0.7076& 0.7921& 0.7094& 0.7960& 0.8295&0.7534 \\
        & FNO+DDELD & 0.9733& 0.9958&0.8688 & 0.9553& 0.9899& 0.9989& 0.9920& 0.9953& 0.9594& 0.9785\\
        & CNN & 0.6262&0.9055 &0.8138 & 0.5901& 0.7933& 0.8791& 0.7877&0.8842 & 0.8972& 0.7943\\
        & CNN+DDELD &0.9301& 0.9669& 0.8628& 0.9443& 0.9729& 0.9807& 0.9651&0.9797 & 0.9817 & 0.9454\\
    \bottomrule   
    \end{tabularx}
    \label{tab:full_data}
\end{sidewaystable}

\subsection{Operator performance with regard to solution frequency behavior}\label{sec:freq_results}
To understand how the localized operator behaves under the DDELD, we perform a sequence of numerical experiments of different solution frequencies as stated in Section~\ref{sec:transport}. We examine the reconstructed $R^2$ accuracy of the solved solution field by FNO in correlation with the decomposed domain size and the character frequency of the solution field and report the result in Figure \ref{fig:multi_f}. 
\begin{figure}
    \centering
    \includegraphics[width=0.8\linewidth]{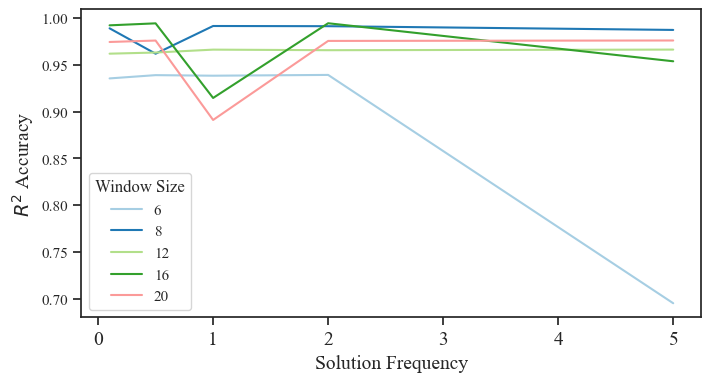}
    \caption{$R^2$ reconstruction accuracy of decomposed model predictions about window size and solution field frequency.}
    \label{fig:multi_f}
\end{figure}
The frequency characterization of the domain reflects the speed at which the information travels in the system. The higher the frequency, the faster the information travels and the larger the local-dependent region of the system becomes. For a fixed window size, as the frequency increases, the local-dependent region will gradually outgrow it and the model will be given less than required information to make the prediction at some point. 
That is why the accuracy drops dramatically as frequency increases for the smallest window size.
It can also be explained in the view of frequency. A domain decomposition is a multiplication between the original solution function and a designated window function. Such multiplication therefore implements a frequency cutoff that adds high-frequency components to the Fourier domain of the original solution function. The higher the frequency of the original solution, the more divergence it will get when restricted to a localized area. 
To alleviate the effects of the frequency cut-off, we can increase the window size. However, as shown in Figure \ref{fig:multi_f}, a larger window size does not necessarily lead to better results. 
We see better solution quality as the decomposition window size increases from 6 to 10, however in high-frequency regions ($f>1$) as the window size gets larger than 12, the averaging accuracy obtained starts to decrease. It could be explained that the increased window size brings information not relevant to the local physical evolution, which is essentially noise so that it hinders the model from capturing the real physical pattern.
It implies that there exists an optimal window size for a specific generic transport system. The results of the experiments indicate that the character frequency of the domain plays an important role in determining the window size.

\subsection{Processing speed analysis}\label{sec:time_cost}
The workhorses of the algorithm are domain decomposition and prediction integration. As analyzed in Sections \ref{sec:data_decomposition} and \ref{sec:prediction_integration}, the time cost of the domain decomposition and window patching algorithms is linear with the largest block number in all dimensions $B_{max}$. A time cost experiment that collects the time cost of the two algorithms under different $B_{max}$ is conducted to demonstrate the linear time cost, as shown in Figure \ref{fig:time_cost}.
\begin{figure}
    \centering
    \includegraphics[width=0.6\linewidth]{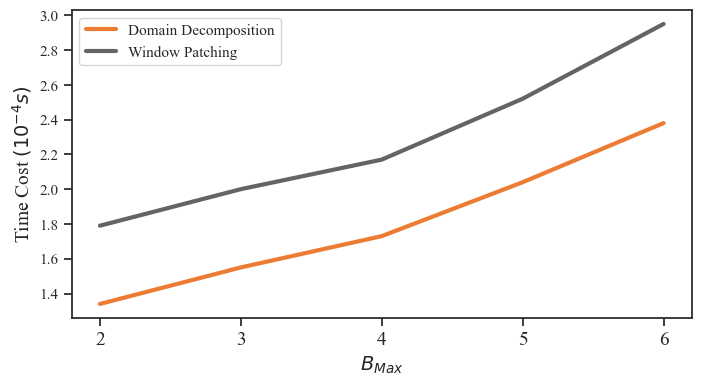}
    \caption{The time cost of the domain decomposition and window patching algorithm is linear with the largest block numbers in all dimensions $O(B_{max})$. For reference, the cost of FNO model inference is $2.1 \times 10^{-3}$ s. The specification of the computer running the time analysis is AMD Ryzen Threadripper PRO 5955WX CPU with 16-Cores, an NVIDIA GeForce RTX 4090 GPU, and 258GiB of memory. }
    \label{fig:time_cost}
\end{figure}

\section{Remarks on Multi-scale Problems}
Many complex systems involve the evolution of physics across multiple scales. For instance, in turbulence modeling, the size of coherent eddies ranges from the large eddy scale that drives turbulence to the Kolmogorov length scale which is determined by viscosity \cite{drikakis2019multiscale}. Given the extremely high computational cost of direct numerical simulation across the full length scales \cite{pope2000turbulent}, multi-scale simulation methods have been developed to reduce this burden in common engineering applications, such as the Reynolds Averaged Navier–Stokes (RANS) equations \cite{durbin2018some}. Our method can support multi-scale problems by training multiple ML models with different window sizes for various physical fields and integrating the results from these models. The mesh-independent properties of certain neural operators, such as the Fourier Neural Operator (FNO), facilitate the merging of subdomains across different scales. The determination of the window size for each physical field follows the same process as for single-scale problems, by specifying a characteristic length and fine-tuning its multiple. 

\section{Conclusion and Future Direction}
In this paper, we reveal the incompatibility between the deep learning architecture and local dependency of time-dependent systems. We prove that the local-dependent region of deep learning models expands inevitably as the number of layers increases. On one hand, the expanded local-dependent region complicates the input data and introduces noise, which detrimentally impacts the convergence rate and generalizability of the models. On the other hand, the expressiveness of the ML models largely relies on the number of layers, so limiting the number of layers would weaken the performance of the models as well. Such a dilemma is caused by the coupled expressiveness and local-dependent property of deep learning architecture. To decouple the two, we propose an efficient data decomposition method. Through the numerical experiments over the data generated by three typical time-dependent PDEs, we analyzed the properties of DDELD (e.g. the relationship among window size, system frequency, and error), and demonstrated its capabilities in accelerating convergence and enhancing generalizability of the ML models. The proposed method has the potential to be extended to unstructured data, like irregular meshes. Our future work will explore the scalability provided by DDELD in parallel computation for complex, large-scale generic transport problems.

\section*{Acknowledgements}
This research is supported by Carnegie Mellon University’s Manufacturing Futures Institute, made possible by the Richard King Mellon Foundation. This material is also based upon work supported by the Engineer Research and Development Center (ERDC) under Contract No. W912HZ22C0022. Any opinions, findings, conclusions, or recommendations expressed in this paper are those of the authors and do not necessarily reflect the views of the sponsors.

\bibliographystyle{unsrt}
\bibliography{ref}

\newpage
\appendix

\section{Proofs}\label{appendix:proof}
We prove Theorem \ref{theorem:local_dependent_region} here.
\begin{lemma}\label{lemma:1}
Given a point $x$ in a metric space $M$ and positive real numbers $\delta_1$ and $\delta_2$, we have   
    $\bigcup_{y\in U(x,\delta_1)} U(y,\delta_2)= U(x, \delta_1+\delta_2)$.
\end{lemma}
\begin{proof}

(1) $\forall z \in \bigcup_{y\in U(x,\delta_1)} U(y,\delta_2)$, $\exists y_0 \in U(x,\delta_1)$, s.t. $z \in U(y_0,\delta_2)$. 
According to triangle inequality, we have 
\begin{equation}
 d(z,x) \leq d(z,y_0) + d(y_0,x).
\end{equation}
Since $d(z,y_0) < \delta_2$ and $d(y_0,x) < \delta_1$, we have
\begin{equation}
     d(z,x)< \delta_1+\delta_2.
\end{equation}
So, $z \in U(x,\delta_1+\delta_2)$.

(2) $\forall z \in U(x, \delta_1+\delta_2)$, if we assume $\nexists y \in U(x,\delta_1)$ s.t. $z\in U(y,\delta_2)$, we show in the following that it will result in contradiction. 
According to the assumption, we have $d(z,y) > \delta_2,\forall y\in U(x,\delta_1)$. 
We denote $\xi = \sup_{z\in U(x, \delta_1+\delta_2)} d(z,x)$. Since $d(z,x)\leq d(z,y)+d(y,z)$, we have
\begin{equation}
    \xi = \sup d(z,y)+\sup d(y,x). 
\end{equation}
Since $\sup d(y,x) = \delta_1$ and $\sup d(z,y) > \delta_2$, we have 
\begin{equation}
  \xi > \delta_1+\delta_2,
\end{equation}
which contradicts with $z \in U(x, \delta_1+\delta_2)$. 
Therefore, $\exists y \in U(x,\delta_1)$ s.t. $z\in U(y,\delta_2)$. So, $z \in \bigcup_{y\in U(x,\delta_1)} U(y,\delta_2)$. 

According to (1) and (2), we have $\bigcup_{y\in U(x,\delta_1)} U(y,\delta_2)= U(x, \delta_1+\delta_2)$.
\end{proof}
Now we can prove Theorem \ref{theorem:local_dependent_region}.
\begin{proof}
Since the non-linear activation function $\sigma$, and the linear project mapping $P$ and $Q$ do not influence the size of the local-dependent region, we can simplify the expression of a neural operator as
\begin{equation}
\begin{aligned}
    & v_{i+1} = K_{\phi}(v_i),\\
    & K_{\phi}(v_i)(x) = \int_{U(x,\delta)} k_{\phi}(x-y)v_i(y)dy.\\
\end{aligned}
\end{equation}

(1) When $k=1$, there is only one layer of local-dependent convolution whose integral is defined over $U(x,\delta)$. So the local-dependent region of $v_1(x)$ is $U(x,\delta)$.

(2) When $k=i$, we assume the local-dependent region of $v_i(x)$ is $U(x,i \delta)$. When $k=i+1$, 
we have
\begin{equation}\label{equ:v_i}
v_{i+1}(x) = \int_{U(x,\delta)} k_{\phi}(x-y)v_i(y)dy. 
\end{equation}
From Equation \ref{equ:v_i}, we know that the calculation of $v_{i+1}(x)$ involves $v_i(y), \forall y\in U(x,\delta)$. So we have the local-dependent region of $v_{i+1}(x)$ as $\bigcup_{y\in U(x,\delta)} U(x, i\delta) = U(x, (i+1)\delta)$ according to Lemma \ref{lemma:1}.
\end{proof}

\newpage
We prove Theorem~\ref{theorem:frequency_based_domain} here.
\begin{lemma}\label{lemma:poisson}
    (Poisson's Summation Formula) If $f\in L_2(\mathbb{R}), B>0$ and 
    \begin{equation}
        \sum_{n\in \mathbb{Z}}\hat{f}(\xi+2Bn)\in L_2([0, 2B]),
    \end{equation}
    then
    \begin{equation}
        \sum_{n\in \mathbb{Z}}\hat{f}(\xi+2Bn)=\frac{1}{2B}\sum_{n\in\mathbb{Z}}2Bf(\frac{n}{2B})e^{\frac{2\pi in\xi}{2B}}
    \end{equation}
\end{lemma}
Proof of Lemma~\ref{lemma:poisson}, or the Poisson's summation formula can be found in \cite{kubota1974analogy}.
\begin{definition}
    (Discrete Fourier Transform) Given a sequence of $N$ complex points $\{f(x_i)|i=1, 2, 3, \cdots, N\}$, the discrete Fourier transform (DFT) of such points into another sequence of complex numbers $\{\hat{f}(x_k)|k=1, 2, 3, \cdots, N\}$ by: 
    \begin{equation}
        \hat{f}(x_k)=\sum_{i=1}^N f(x_i)\cdot e^{-2\pi j\frac{k}{N}i}.
    \end{equation}
\end{definition}
We can then prove Theorem~\ref{theorem:frequency_based_domain}:
\begin{proof}
    Since $\text{supp}(\hat{f})\subseteq [-B, B]$, according to Lemma~\ref{lemma:poisson} we can obtain a maximum bound for $f(x)$ as: 
    \begin{equation}
        \sum_{n\in\mathbb{Z}}f\left(x+\frac{n}{2B}\right)\in L_2\left(\left[0, \frac{1}{2B}\right]\right).
        \label{eqn:f1}
    \end{equation}
    Suppose we take a unit length of a total of $N$ points (which, for simplicity, we assume $N$ can be divided by 2 though it doesn't have to) in $n\in\mathbb{Z}$ Equation~\ref{eqn:f1} also holds:
    \begin{equation}
        \sum_{n=-N/2}^{N/2}f\left(x+\frac{n}{2B}\right)\in L_2\left(\left[0, \frac{1}{2B}\right]\right).
        \label{eqn:f2}
    \end{equation}
    We can then write the Fourier series of Equation.~\ref{eqn:f2} as:
    \begin{equation}
        \sum_{n=-N/2}^{N/2}f\left(x+\frac{n}{2B}\right)=\sum_{m\in\mathbb{Z}}c_m e^{2\pi jmx\cdot 2B},
    \end{equation}
    where,
    \begin{equation}
    \begin{split}
        c_m&=2B\int_0^{\frac{1}{2B}}\sum_{n=-N/2}^{N/2}f\left(x+\frac{n}{2B}\right)e^{-2\pi jmx\cdot 2B}\text{d}x,\\
        &=\sum_{n=-\frac{N}{2}}^{\frac{N}{2}}2B\int_0^{\frac{1}{2B}}f\left(x+\frac{n}{2B}\right)e^{-2\pi jmx\cdot 2B}\text{d}x.
    \end{split}
    \label{eqn:cm}
    \end{equation}
    Substitute new variable $y=x+\frac{2B}{n}$ into the RHS of Equation~\ref{eqn:cm} we get:
    \begin{equation}
    \label{eqn:pre-monte}
    \begin{split}
        c_m&=\sum_{n=-\frac{N}{2}}^{\frac{N}{2}}2B\int_{\frac{n}{2B}}^{\frac{n+1}{2B}}f(y)e^{-2\pi jmy\cdot 2B}\text{d}y\\
        &=2B\int_{\frac{-N}{4B}}^{\frac{N+2}{4B}}f(y)e^{-2\pi jmy\cdot 2B}\text{d}y.
    \end{split}
    \end{equation}
    Since we can only obtain point-wise observations of the domain, we replace the integration in Equation~\ref{eqn:pre-monte} with a summation, and for per unit length obtain:
    \begin{equation}
    \label{eqn:aft-monte}
        c_m=2B\sum_{y=\frac{-N}{4B}}^{\frac{N+2}{4B}}f(y)e^{-2\pi jmy\cdot 2B}.
    \end{equation}
    Notice the similarity between Equation~\ref{eqn:aft-monte} and the formulation of DFT on a set of numbers $\{f(y)|y\subseteq[\frac{-N}{4B}, \frac{N+2}{4B}]\}$. We can conclude that, To fully recover the frequency character of $f$, the number of nodes to include in the localized region must satisfy:
    \begin{equation}
        L_c=\lceil\frac{N+2}{4B}+\frac{N}{4B}\rceil=\lceil\frac{N+1}{2B}\rceil,
    \end{equation}
    Where $B$ is the frequency bandwidth of the domain, and $N$ is the total number of nodes per unit length.
\end{proof}

\paragraph{Remark}
It is worth noting that theorem~\ref{theorem:frequency_based_domain} is essentially a variance of the Nyquist-Shannon sampling theorem~\cite{jerri1977shannon} conditioned by the domain's size of discretization. In fact, we borrowed many tools from Shannon's original proof during the process of reaching theorem~\ref{theorem:frequency_based_domain}. This interesting similarity between signal processing methods and physics domain selection further consolidates the argument that machine learning methods in solving PDEs are far more sensitive to the initial condition and the structure of physical properties in the field rather than specific PDE coefficients and that the change in initial conditions can have a big impact on the performance of machine learning methods even when the PDE itself remains the same. This frequency-based analysis also sheds light on applying the DDELD to other frequency-dominant scenarios such as isotropic turbulence, and we hope to address such issues in future work.

\end{document}